\pdfoutput=1

\documentclass[11pt]{article}

\usepackage[final]{acl}

\usepackage{times}
\usepackage{latexsym}
\usepackage{natbib}
\usepackage{amsmath}
\usepackage{booktabs}
\usepackage{tabularx}
\newcolumntype{b}{X}
\newcolumntype{s}{>{\hsize=.15\hsize}X}
\newcolumntype{m}{>{\hsize=.5\hsize}X}
\usepackage{pifont}
\usepackage{enumitem}
\usepackage{listings}
\usepackage{pdfpages}
\usepackage[T1]{fontenc}

\usepackage[utf8]{inputenc}

\usepackage{microtype}

\usepackage{inconsolata}

\usepackage{graphicx}
\usepackage{xcolor}

\title{Bringing Pedagogy into Focus \includegraphics[scale=0.006]{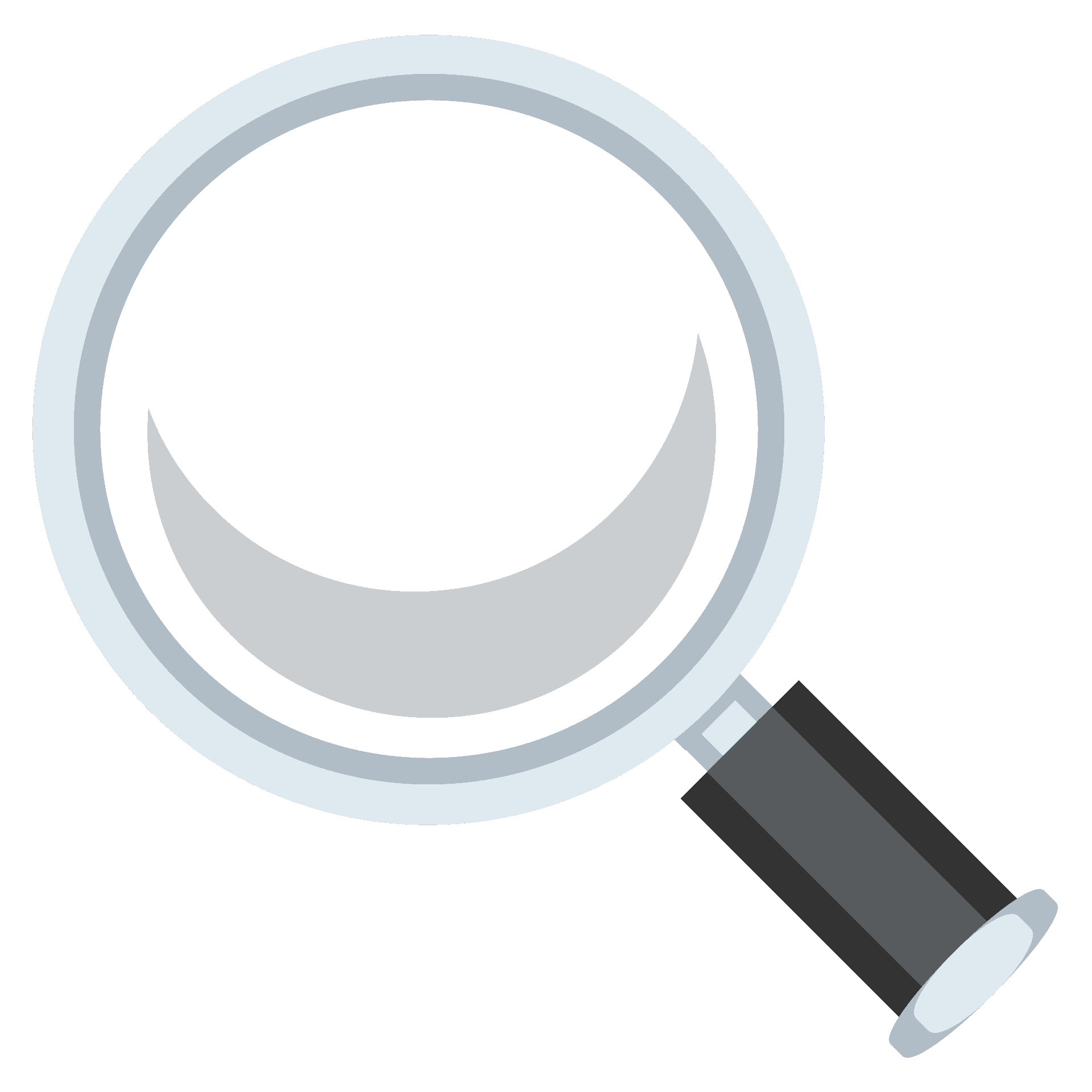}: Evaluating Virtual Teaching Assistants’ Question-Answering in Asynchronous Learning Environments}

\author{
 \textbf{Li Siyan}\thanks{Co-first authors},
 \textbf{Zhen Xu}\footnotemark[1],
 \textbf{Vethavikashini Chithrra Raghuram},
 \textbf{Xuanming Zhang},\\
 \textbf{Renzhe Yu}\thanks{Co-senior authors},
  \textbf{Zhou Yu}\footnotemark[2]
\\
 Columbia University
\\
 \texttt{\href{mailto:siyan.li@columbia.edu}{siyan.li@columbia.edu},\href{mailto:zx2393@tc.columbia.edu}{zx2393@tc.columbia.edu}
 }
}

\begin{document}
\maketitle
\begin{abstract}
Asynchronous learning environments (ALEs) are widely adopted for formal and informal learning, but timely and personalized support is often limited. In this context, Virtual Teaching Assistants (VTAs) can potentially reduce the workload of instructors, but rigorous and pedagogically sound evaluation is essential. Existing assessments often rely on surface-level metrics and lack sufficient grounding in educational theories, making it difficult to meaningfully compare the pedagogical effectiveness of different VTA systems. To bridge this gap, we propose an evaluation framework rooted in learning sciences and tailored to asynchronous forum discussions, a common VTA deployment context in ALE. We construct classifiers using expert annotations of VTA responses on a diverse set of forum posts. We evaluate the effectiveness of our classifiers, identifying approaches that improve accuracy as well as challenges that hinder generalization. Our work establishes a foundation for theory-driven evaluation of VTA systems, paving the way for more pedagogically effective AI in education.
\end{abstract}

\section{Introduction}

Asynchronous learning environments (ALEs) have become a critical complement and extension to traditional synchronous learning (e.g., live lectures and real-time tutoring). By removing time and location constraints, ALEs enhance flexibility and inclusivity, supporting more equitable access to educational resources and opportunities \cite{tate2022equity,farley2022online}. However, providing timely and effective instructional support remains one of the biggest challenges for ALE adoption. Their self-paced, decentralized nature makes it difficult for instructors to track progress and manage interactions, often leading to delayed feedback that can reduce student motivation, engagement, and learning effectiveness \cite{lin2020students,watts2016synchronous}.

Virtual Teaching Assistants (VTAs) offer a promising approach to address this challenge by providing scalable, timely, and personalized support, helping reduce instructor workload in ALEs \cite{audras2022virtual}. They are often powered by Retrieval-Augmented Generation (RAG) approaches to respond quickly to student questions \cite{liu2024step}. While earlier systems focused on routine and syllabus-based questions \cite{goel2018jill, sajja2023platform, raamadhurai2019curio}, recent VTAs powered by Large Language Models (LLMs) can tackle more complex academic queries \cite{taneja2024jill,liu2024step, laney2024human, qiao2025oversight}. 
Although these advancements highlight the growing capabilities of VTAs, the pedagogical effectiveness of their generated content still lacks a systematic and scalable evaluation framework, which is a critical issue. Early approaches often rely on manual qualitative analyses \cite{lee2023learning, li2023curriculum} or general-purpose language metrics \cite{al2024can, vasselli2023naisteacher} to assess similarity to human responses. Nevertheless, these metrics are not designed to assess teaching quality and often lack generalizability across different educational contexts.

To bridge this gap, we propose a \textbf{pedagogical teaching assistant evaluation framework grounded in learning science}, focusing on VTAs in asynchronous online forums, one of the most common deployment contexts in ALEs. The framework has two components: a five-level pedagogical skill model with progressively increasing ``instructional difficulty'': \textit{(1) Clarify Misunderstandings and Confusions}, \textit{(2) Deepen Disciplinary Understanding}, \textit{(3) Foster Higher-Order Thinking}, \textit{(4) Enhance Metacognitive Awareness}, and \textit{(5) Foster Collaborative Knowledge Construction} developed by linking core cognitive mechanisms to pedagogical goals, and an evaluation rubric that operationalizes this model by specifying observable instructional behaviors at each level.

When establishing the usability and reliability of our rubric, we utilize a set of simulated VTA responses on student discussion forum posts collected through an IRB-approved procedure. Note that Level 5 of our framework is currently unattainable for most VTAs, as their responses are typically limited to a single forum post and any retrieved content, excluding broader context from other posts in the discussion. However, asynchronous online forums are inherently community-driven spaces, and achieving collaborative knowledge construction requires access to broader, forum-level context. 
To simulate VTAs with broader forum-level post access for Level 5 evaluations, we use RAG to create responses informed by similar peer posts. Expert annotators apply our rubrics to these simulated responses. We then leverage these annotations to build automatic classifiers using LLMs.

\citet{maurya2024unifying} reports negative correlations between LLM and expert annotations in tutor pedagogical assessment, indicating that LLMs may be unreliable evaluators. Their study uses a single, unoptimized prompt without fine-tuning, which the authors acknowledge as a limitation. 
We extend this line of inquiry to the VTA pedagogical evaluation context by systematically exploring whether prompt optimization and fine-tuning can improve alignment between LLM-based evaluators and expert pedagogical judgments.

Our expert annotation results indicate that \textbf{including peer forum posts reduces pedagogical capability in Levels 1-4} within our specific simulated VTA environment.
Furthermore, while our automated classifiers perform reliably on lower-level, in-distribution examples, they \textbf{struggle at higher pedagogical levels} and \textbf{exhibit limited generalization}, even for Level 1. 

Overall, our contributions are as follows:
\begin{enumerate}
    \item We introduce the first pedagogical evaluation framework designed to be generalizable within the context of asynchronous online forums, grounded in learning science theory.
    \item We demonstrate the generalizability of our evaluation framework by applying it to both open-source and synthesized VTA responses.
    \item We show that current LLMs struggle with our nuanced classification task, highlighting the need for caution when applying LLMs to pedagogical evaluation.
\end{enumerate}

\section{Related Work}

\subsection{Teaching System Evaluation Frameworks}

Pedagogical skills are complex and context-dependent, making automated evaluation difficult. Early work in evaluating VTA relied on manual qualitative analysis, usage logs, measures of learning outcomes, or generic language metrics \cite{wang2023bridging, lee2023learning, li2023curriculum, kweon2025large, zamfirescu202561a, al2024can, vasselli2023naisteacher,dan2023educhat, dahal2025autota}, which lack specificity for educational contexts. Recent studies have proposed more targeted frameworks for pedagogy automated evaluation. For example, \citet{demszky2021measuring} introduced ``uptake'' to capture how instructors build on student input, while \citet{tack2022ai} outlined three pedagogical dimensions: teaching alignment, student understanding, and helpfulness. \citet{maurya2024unifying} expanded this with a taxonomy of eight learning science–grounded dimensions, and \citet{jurenka2024towards} proposed seven diverse educational benchmarks from high-level learning science principles, including an LLM-based automatic evaluation method. Their evaluation framework includes five teaching dimensions and eight metrics. However, these efforts focus on synchronous tutoring settings, whereas asynchronous forums emphasize timely feedback, critical thinking, and peer dialogue. Despite the growing use of VTAs in such forums, evaluation still relies predominantly on student surveys, qualitative content analysis of generated responses, and usage log–based behavioral metrics \cite{goel2018jill,hicke2023ai,baker2024step,sinha2024boilertai,qiao2025oversight}, with no pedagogical evaluation framework specifically tailored to this context.

\subsection{LLMs as Evaluators for Education}

Using LLMs as evaluators has become increasingly popular \cite{chang2024survey,chan2023chateval}, particularly within educational contexts. Recent studies have employed LLMs to evaluate the quality of student writing \cite{zhang2024decor,kim2024can,zhang2024prolex} and to judge the human-likeness of AI tutor responses \cite{chevalier2024language}. However, their reliability in assessing complex constructs, such as pedagogical capabilities, remains unclear \cite{maurya2024unifying,wang2023large}. 
A key challenge is that general-purpose evaluation methods could fail to align with the complex, ambiguous, and domain-expert inferences required by many educational assessments \cite{han2023llm,xu2025course,wei2025elmes}. Research has shown that LLM evaluation performance can vary across task context and design \cite{liu2025qualitative}. These disparities underscore the need for further investigation into LLM-based educational evaluation, informed by domain expertise, to avoid overestimating what general-purpose LLM evaluators can accurately and reliably achieve.



\section{Our Evaluation Framework}
Evaluating pedagogical skills is challenging because teaching support is a dynamic process that varies by context, subject, and learner preferences. Therefore, effective evaluation requires selecting the appropriate level of observation granularity and ensuring alignment with the specific learning context. In our framework, we follow three steps to develop the pedagogical skill model and the evaluation rubric:(1) identifying cognitive mechanisms in asynchronous online forums, (2) distilling context-specific pedagogical goals, and (3) translating these goals into observable pedagogical behaviors. 

\subsection{Mapping Cognitive Mechanisms}
Asynchronous online forums are typically structured by instructors through predefined discussion prompts, allowing students to engage at their own pace by asking questions, sharing ideas, and interacting with peers. Compared to synchronous instruction or other forms of asynchronous learning, such as lectures or tutoring, the key feature of online discussion forums is their decentralized, learner-driven, and socially interactive nature. Rather than following a predefined or instructor-led sequence, these forums rely on students to initiate, respond to, and sustain the discourse. The content is also more open-ended and exploratory, with fewer structural constraints.

From a cognitive science perspective, learning in asynchronous online forums is grounded in constructivist and Vygotsky’s sociocultural learning theories \cite{fosnot2013constructivism, lantolf2000introducing}, which view learning as an active, socially mediated process of meaning-making. Several cognitive mechanisms support effective learning in this context: clarifying confusion and constructing understanding through questioning, exposing misconceptions, and receiving explanatory feedback \cite{hew2010student}; fostering higher-order thinking through advanced cognitive processes such as analysis, reasoning, and argumentation \cite{meyer2003face, wang2016towards}; promoting reflection and self-regulation through self-explanation and elaboration \cite{choi2005scaffolding}; and deepening disciplinary understanding through exposure to diverse perspectives and information. Interaction with conflicting ideas can also trigger cognitive conflict, prompting learners to identify gaps in their thinking, facilitate conceptual change, and integrate new knowledge into existing cognitive schema \cite{limon2001cognitive}.

\subsection{Defining Pedagogical Goals}
\label{sec:pedagogical_principles}


Building on the learning characteristics and cognitive mechanisms summarized in the previous section, we define the instructional support goals for asynchronous online forums and organize them hierarchically by cognitive engagement and instructional complexity, in accordance with well-established theories (ICAP)\cite{chi2014icap} of cognitive engagement in the learning sciences.
\begin{enumerate}
    \item \textbf{Clarify Misunderstandings and Confusions:} Identify underlying misunderstandings and address learner confusion by providing clear, instructional explanations to enhance conceptual clarity \cite{stahl2013model,dykman2008part,nandi2012evaluating,hew2010student,pena2004analyzing}.
    \item \textbf{Deepen Disciplinary Understanding:} Deepening students’ understanding of disciplinary concepts through appropriate and effective instructional support \cite{engle2002guiding,yang2022fostering,onyema2019online}.
    \item \textbf{Foster Higher-Order Thinking:} Stimulates students’ engagement in complex cognitive processes and the application of higher-order thinking skills \cite{mcloughlin2009analysis,wang2016towards,szabo2011learning,brierton2016comparison,yang2005using,onyema2019online}.
    \item \textbf{Enhance Metacognitive Awareness:} Supports students’ reflection on their understanding, recognition of knowledge gaps, and evaluation of their thinking and learning strategies \cite{murphy2009framework, michalsky2007developing,akyol2011assessing}.
    \item \textbf{Foster Collaborative Knowledge Construction:} Encourages meaningful peer interaction, supports the exchange of diverse perspectives, and fosters a collaborative learning environment and sense of community \cite{chi2014icap,zou2021exploring,akcaoglu2016increasing,joksimovic2015social}.
\end{enumerate}

\begin{table*}[!h]
    \centering
    \begin{tabularx}{\textwidth}{mmb}
    \toprule
       \textbf{Course Information} & \textbf{Discussion Topic} & \textbf{Forum Post Content}\\
       \midrule
        \textit{PHYSICS 103. Intro. to Astronomy.} \newline The history of Astronomy as a field, early... & Question Week 6
        Use this message board to only post questions relevant to course...& can someone explain how redshift works and what it tells us about galaxies moving away from us? also, how do exoplanets get detected with the transit method? appreciate any explanations!\\
        \midrule
         \textit{MGMT 157. Taxation of Individuals.} The fundamentals of federal... & INTRO - Dependents: Items from Dependents Module ... & If two individuals are both helping support someone, like a parent, how to figure out who gets to claim the parent?\\
    \bottomrule
    \end{tabularx}
    \caption{Mock example student discussion forum posts in our dataset. These examples are created to be similar to the real examples to preserve student privacy.}
    \label{tab:example_posts}
\end{table*}

\subsection{Developing Rubric of Observable Behaviors}

Based on the pedagogical goals outlined above, we develop an evaluation rubric to capture distinct pedagogical capabilities within each level. The rubric construction process includes three iterative stages to ensure both theoretical alignment and empirical validity.

In the first stage, we drew upon prior research on effective instructional practice in online forums to draft initial level descriptions. These descriptions operationalized pedagogical behaviors at varying degrees of effectiveness for each level. In the second stage, we refined the rubric by conducting interviews with two experienced instructors, each with over three years of experience facilitating discussions. This empirical review enabled us to revise and align the rubric more closely with authentic instructional practices. In the final stage, we conduct a pilot validation by recruiting two graduate students in education to independently apply the draft rubric to a sample of 20 forum responses. They then discuss any discrepancies and collaborate on refining the rubric descriptions to ensure that all levels are clearly distinguishable and appropriately represented. The final rubric, along with illustrating examples, is in Appendix \ref{sec:rubrics}.
Each pedagogical level has possible ratings of 0, 1, 2, or ``NA''.

\section{Pedagogical Evaluation of Simulated VTA Responses}
We apply our evaluation framework to two sets of simulated VTA responses, one set with similar peer posts and the other without. This is to examine if incorporating peer responses as forum-level context affects the expressed pedagogical capabilities of our LLM-powered VTA. 

\subsection{Source of Student Discussion Forum Posts}

To simulate VTA responses, we use a dataset of real online forum posts in thousands of courses offered at a large public university in the United States, available through our existing research partnership. IRB approval and data use agreements have been in place for the project. Complying with the Family Educational Rights and Privacy Act (FERPA), all personally identifiable information in the raw data was removed or redacted prior to researcher access, and the dataset cannot be made publicly available. Nonetheless, we release our code for data generation and classification at \url{https://github.com/Columbia-NLP-Lab/VTA_qa_evaluation}.

We construct a classifier using \texttt{GPT-4o-mini} to categorize discussion forum posts into five categories, one of which is academic questions (see more details about this simple classifier in Appendix \ref{app:dspy_question_classifier}).
While our pedagogical framework is designed to be generalizable across various forms of student engagement, we selected academic questions as our starting point in this work, as this allows us to assess the foundational pedagogical capabilities of VTAs. Note that we aim to identify academic questions with high precision. Thus, the overall question classification accuracy is not our primary concern. The original set of data consists of 600k forum posts, from which we select the first 60k for classification. Additionally, we filtered for posts that are the beginning of their post threads.

For this study, we focus on a diverse sample set of 300 academic questions from 116 students. These questions offer high coverage of academic disciplines, spanning 85 courses and 40 departments, from criminal law to transportation systems.


\begin{figure*}[!h]
    \includegraphics[width=0.9\textwidth]{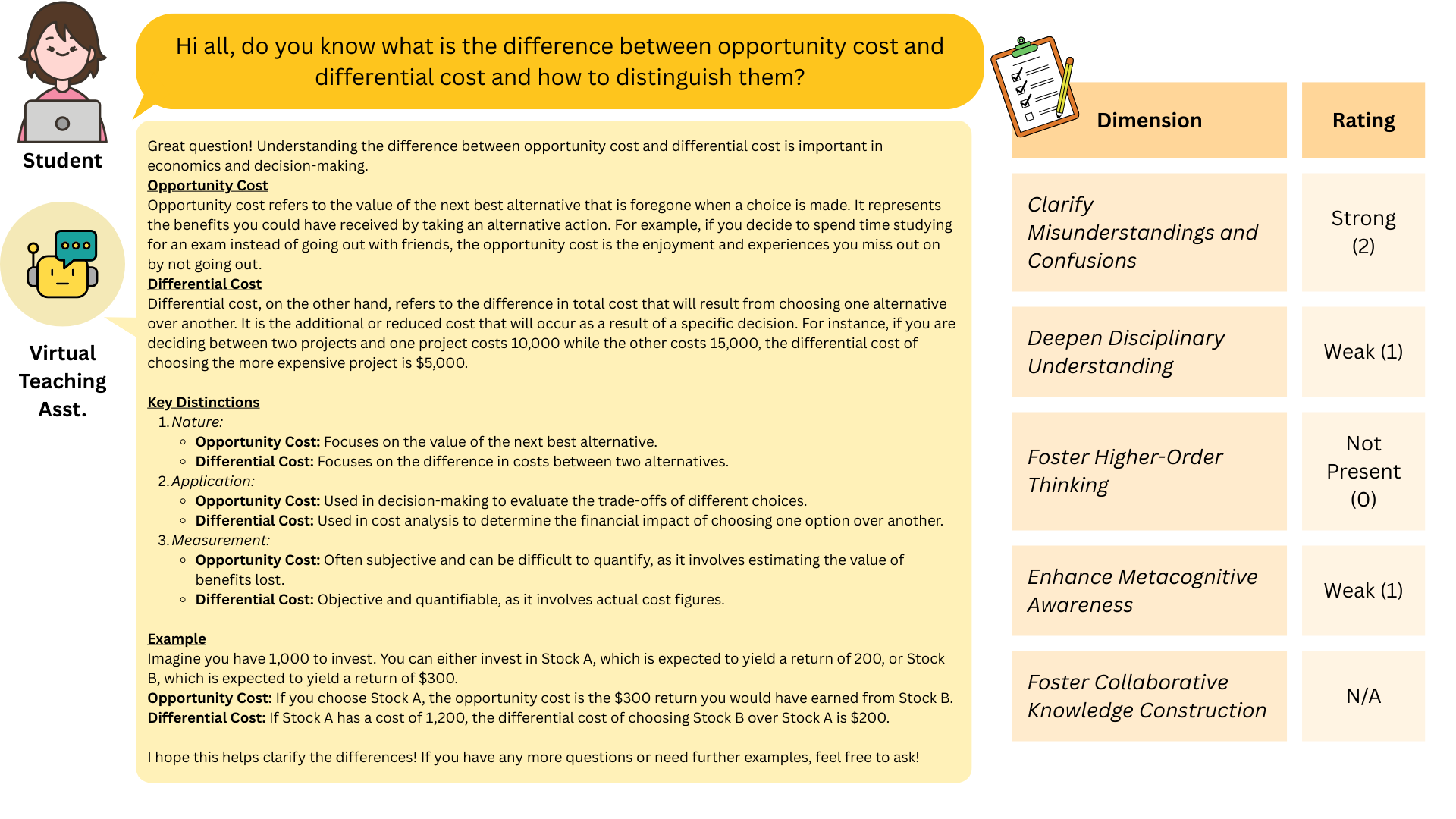}
    \caption{A mock student post coupled with a VTA response generated using the system from \citet{kweon2025large}, rated along the pedagogical dimensions from our framework.}
    \label{fig:example_vta_gen}
\end{figure*}

\subsection{Example VTA Response Generation}
\label{sec:vta_gen_process}

We select \texttt{Llama-3-70B-Instruct} \cite{grattafiori2024llama} as our primary LLM to generate simulated VTA responses because we want instances of pedagogical attempts with varying degrees of success according to our framework. Due to the lack of access to course materials, we prompt the model to respond to student questions using its inherent knowledge base instead of taking retrieval-augmented generation approaches. 

Two sets of responses are generated: with and without forum-level context.

\noindent{\textbf{Context-Free:}} For each academic question, the default VTA prompt includes: (1) Course information, when available. We scraped the course catalog from the university to obtain course names and descriptions. (2) Description of the discussion topic. (3) Content of the student's forum post. Mock example data instances are in Table \ref{tab:example_posts}. 

We rewrite the first four levels of our pedagogical framework into goals and incorporate them into the default prompt to define desirable behaviors. We exclude the fifth level here, since there is no peer information available. 

\noindent{\textbf{Forum-Level Context:}} To provide this context, we retrieve the top ten most similar student posts (excluding the post to be responded to) using OpenAI's \texttt{text-embedding-3-small} embedding. When there is a discussion topic associated with the forum post, we select contextual forum posts under the same discussion topic; otherwise, from the entire course. All relevant posts are labeled with consecutive numbers, so that the VTA can refer to the posts as "Similar Post \#1", etc. 

We include the goal-style rewrite of Level 5 in this prompt version. See the prompts in Appendix \ref{sec:vta_prompts} and example generations in Appendix \ref{sec:example_gens}. We focus on simulated VTA responses in this work to ensure comparison fairness between the different generation conditions. It is important to note, however, that our framework is already applicable to current VTA systems. We include an example in Figure \ref{fig:example_vta_gen} with a mock student discussion forum post, the corresponding VTA response, and the assessments using our framework.






\subsection{Annotation Procedure}\label{subsec:annotation}

Given the pedagogical knowledge and background required for accurate ratings, we recruit two volunteer annotators who are pursuing graduate degrees in education, with specializations in learning sciences and pedagogy. Before the annotation, one of the authors conducts a training session with the two annotators to clarify the coding rubric and the definitions of each rating level. During the session, the annotators independently label a set of 20 sample VTA responses and discuss any disagreements or ambiguities in their interpretations. The rubric descriptions are revised to enhance clarity for subsequent annotations.

Then, two annotators independently rate the full set of VTA responses. At the 80-response milestone, the author conducts an inter-rater agreement check, during which the annotators achieve an intraclass correlation coefficient (ICC) of 0.81, indicating a fine agreement and minimal misunderstanding. Upon completion of the full annotation, the final ICC was 0.78, indicating good reliability \cite{koo2016guideline}.

Next, we compare the annotations from two annotators. We first identify items with a rating discrepancy greater than 1 (e.g., one annotator rates an item as 0 while the other rates it as 2, suggesting potentially substantive disagreement), which occurs in 1.8\% of our sample. For these cases, one of the authors discusses with both annotators, and the final score for each item is determined by majority agreement. Items with a rating discrepancy of exactly 1, likely reflecting random rating variance, occur in 11.2\% of our sample. These cases are divided equally among the author and the two annotators for a second round of review to determine the final score. The distribution of the annotated scores across each level can be found in Table \ref{tab:score with context} and Table \ref{tab:score without context} in the Appendix.

\section{Automated Pedagogical Evaluations of VTAs}

We explore automatic approaches for pedagogical assessment using LLMs to address practical challenges in using expert annotators and to investigate the claim from \citet{maurya2024unifying} about the limitations of LLM evaluators in education.

The labeling process in Section~\ref{subsec:annotation} yielded 600 unique annotated post-response pairs. We randomly sample 150 context-free and 150 contextual pairs as our training set, consisting of student forum post information, the corresponding VTA responses generated by \texttt{Llama-3}, and ratings on all five levels. Note that the annotators do not label the context-free instances for Level 5; as a result, all levels but Level 5 have 300 training pairs. The rest of the data constitutes a held-out test set. We refer to the training set as \texttt{Llama3-Train} and the test set as \texttt{Llama3-Test} for the rest of the paper.

\subsection{Prompt-based Classifiers}
We experiment with \texttt{GPT-4o-mini} and \texttt{GPT-4o} for zero-shot and prompt-optimized classification. We implement the classifiers as language model programs using the DSPy framework \cite{khattab2023dspy} for compatibility with prompt optimization (Appendix \ref{sec:dspy_program_classifier}). The classifier program receives (1) the course information, (2) the discussion topic, (3) the forum post, (4) the VTA response, and (5) the rubric for \textbf{a specific pedagogical level}, and then \textbf{outputs the scoring for that specific level}.

For prompt optimization, we leverage DSPy's SIMBA prompt optimizer\footnote{\url{https://github.com/stanfordnlp/dspy/blob/main/dspy/teleprompt/simba.py}}. SIMBA stands for Stochastic Introspective Mini-Batch Ascent. This prompt optimizer accepts arbitrary DSPy programs and proceeds in a sequence of mini-batches, seeking to craft incremental improvements as general rules or few-shot examples to the prompts. Improvements are measured using custom metrics, and in our case, we employ classification accuracy.


Prompt optimization yields significant performance gains even with as few as 150 examples. This efficiency allowed us to experiment with two strategies on the \texttt{Llama3-Train} dataset when optimizing prompts: (1) using all available annotations per post-response pair (four to five instances), and (2) using only annotations for specific pedagogical levels (one instance per pair).

\begin{figure*}[t]  
    \centering
    \includegraphics[width=0.9\textwidth]{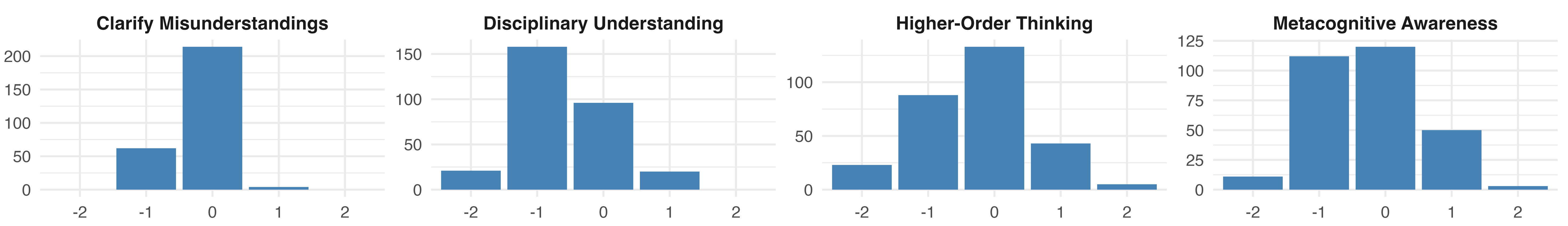}
    \caption{Distribution of score differences (with context $-$ without context) across four pedagogical levels.}
    \label{fig:score differences}
\end{figure*}

\subsection{Open-Weight Classifiers}

To alleviate potential privacy concerns when using API-based frontier LLMs to classify forum posts, we explore open-weight models hosted locally. Unfortunately, our human-annotated dataset is not sufficient to fine-tune these models. Therefore, we perform data synthesis using \texttt{Llama3-Train}.

We use \texttt{GPT-4.1-nano} to generate pairs of student forum posts and VTA responses, each aligned with a specific pedagogical level score. To ensure class distribution balance, we aim to have 300 examples for every score within each pedagogical level. Rubrics are used to define desired behavioral characteristics of VTA responses for each level-score combination. We include both the requirements and sampled in-context examples for generation (detailed description in Appendix \ref{sec:synth_data}).

The above process yields approximately 1200 total examples (synthetic and non-synthetic) per pedagogical level. We split these examples into training and validation sets using an 85:15 ratio. 

We fine-tune \texttt{Qwen3-4B}, \texttt{Qwen3-8B} \cite{yang2025qwen3}, and \texttt{ModernBERT-base} \cite{warner2024smarter} on synthetic data (detailed in Appendix \ref{sec: model_training}), training \textbf{one classifier per pedagogical level} due to the limited generalization of smaller models. Inputs include (1) the student forum post, (2) the VTA response, and (3) the level-specific rubric. Qwen models directly output their ratings, while ModernBERT performs multi-class classification.

\subsection{Validation}

Both prompt-based and open-weight classifiers are evaluated using \texttt{Llama3-Test}. Additionally, to assess generalizability across different LLM-powered VTAs and forum datasets, we synthesize additional test sets, \texttt{New-LLM-Test} and \texttt{Llama3-MOOC}.


\noindent \textbf{Synthesizing \texttt{New-LLM-Test}:} For evaluating cross-model generalizability, we employ the same VTA response generation scheme as detailed in Section \ref{sec:vta_gen_process}, using context-free and contextual prompts. \texttt{GPT-4o}, \texttt{GPT-4.1}, and \texttt{Qwen3-32B} are used to generate six sets of 30 responses (three models, 30 randomly sampled academic questions, and context-free and contextual variants).

\noindent \textbf{Synthesizing \texttt{Llama3-MOOC}:} We leverage the Stanford MOOC dataset, a collection of 30,000 human-annotated MOOC forum posts \cite{agrawal2015youedu}. One of the authors manually went through the student posts labeled as questions and identified 104 academic questions among them. We generate VTA responses using \texttt{Llama-3-70B-Instruct}. See the prompt for this in Appendix \ref{sec:vta_prompts}.

\begin{table*}[!ht]
    \centering
    \begin{tabular}{l|cccccccccc|cc}
    \toprule
    \textbf{Classifier} & \multicolumn{2}{c}{\textbf{Level 1}} & \multicolumn{2}{c}{\textbf{Level 2}} & \multicolumn{2}{c}{\textbf{Level 3}} & \multicolumn{2}{c}{\textbf{Level 4}} & \multicolumn{2}{c}{\textbf{Level 5}} & \multicolumn{2}{c}{\textbf{Average}} \\
     & F-1 & Acc. & F-1 & Acc. & F-1 & Acc. & F-1 & Acc. & F-1 & Acc. & F-1 & Acc.\\
     \midrule
     GPT-4o-mini & \textbf{85.6} & \textbf{87.5} & 52.9 & 61.9 & 40.8 & 48.8 & 49.8 & 50.0 & 32.7 & 38.7 & 52.4 & 57.4 \\
     GPT-4o-mini-Gen. & 84.0 & 85.5 & 53.4 & 55.2 & 42.8 & 48.8 & 53.6 & 54.3 & 37.5 & 42.7 &54.3 & 57.3 \\
     GPT-4o-mini-Lvl. & -- & -- & \textbf{62.6} & \textbf{64.9} & \textbf{53.5} & \textbf{57.6} & 57.0 & \textbf{61.7} & 42.8 & \textbf{49.3} & \textbf{60.3} & \textbf{64.2} \\
     \midrule
     GPT-4o & 84.2 & 84.5 & 55.9 & 57.5 & 47.1 & 47.1 & 30.4 & 31.3 & 41.8 & 41.3 & 51.9 & 52.3 \\
     GPT-4o-Gen. & 85.1 & 84.5 & 48.4 & 47.8 & 48.6 & 50.2 & 56.3 & 55.0 & \textbf{44.5} & 46.0  & 56.6& 56.7\\
     GPT-4o-Lvl. & 83.7 & 86.8 & 52.7 & 53.9 & 47.2 & 50.2 & \textbf{58.8} & 58.3 & 38.7 & 48.0 & 56.2 & 59.4 \\
    \bottomrule
    \end{tabular}
    \caption{Performance of prompting-based classifiers on \texttt{Llama3-Test}. Scores are in percentages. F-1 scores are weighted. Bolded values indicate the best performance in each column.}
    \label{tab:llm_po_results}
\end{table*}

\begin{table*}[!ht]
    \centering
    \begin{tabular}{l|cccccccccc|cc}
    \toprule
    \textbf{Classifier} & \multicolumn{2}{c}{\textbf{Level 1}} & \multicolumn{2}{c}{\textbf{Level 2}} & \multicolumn{2}{c}{\textbf{Level 3}} & \multicolumn{2}{c}{\textbf{Level 4}} & \multicolumn{2}{c}{\textbf{Level 5}} & \multicolumn{2}{c}{\textbf{Average}}\\
    & F-1 & Acc. & F-1 & Acc. & F-1 & Acc. & F-1 & Acc. & F-1 & Acc. & F-1 & Acc.\\
    \midrule
    Qwen3-4B & 83.1 & 87.5 & 50.2 & 63.6 & 13.3 & 21.3 & 34.9 & 49.5 & 17.0 & 33.3 & 39.7 & 51.0\\
    FT-Qwen3-4B & 83.6 & 86.5 & 54.4 & 63.6 & 55.0 & 53.7 & 45.0 & 52.2 & 35.3 & 40.7 & 54.7 & 59.3\\
    \midrule
    Qwen3-8B & 83.6 & 86.5 & 51.3 & 63.2 & 48.9 & 48.3 & 40.4 & 50.5 & 32.0 & 39.3 & 51.2 & 57.6 \\
    FT-Qwen3-8B & \textbf{84.2} & \textbf{86.8} & 59.6 & \textbf{64.2} & \textbf{65.0} & \textbf{69.7} & 45.3 & 51.5 & 35.9 & 40.7 & 58.0 & 62.6\\
    \midrule
    FT-ModernBERT & 81.9 & 84.8 & \textbf{62.4} & 63.2 & 46.8 & 49.2 & \textbf{59.8} & \textbf{68.0} & \textbf{40.0} & \textbf{49.3} & \textbf{58.2} & \textbf{62.9}\\
    \bottomrule
    \end{tabular}
    \caption{Performance of open-weight, fine-tuned classifiers on \texttt{Llama3-Test}. Rows without the ``FT'' prefix represent zero-shot classification results.}
    \label{tab:open_results}
\end{table*}

\section{Results}

\subsection{Human Annotation}

To examine whether providing contextual information influences the pedagogical quality of generated responses from \texttt{Llama-3-70B-Instruct}, we compare expert annotation scores between the two conditions. 
Figure \ref{fig:score differences} records the score differences.

\textbf{Disciplinary understanding (Level 2) experiences the steepest decline with context}, with 59.7\% of responses decreasing by 1 or 2 points. \textbf{Similar patterns are observed in the levels of higher-order thinking (Level 3) and metacognitive awareness (Level 4)}, which show declines of 37\% and 41\%, respectively. We observe that context-free responses focus more on deep engagement with the original post, extending its ideas effectively. In contrast, adding forum-level context often shifts \texttt{Llama-3-70B-Instruct}'s focus to linking similar posts, resulting in broader but more superficial instructional guidance. 



The level of clarification of misunderstandings (Level 1) appears to be the least affected by the inclusion of context. 
Qualitative analysis suggests that context has little impact on how well a response addresses confusion. Instead, the effectiveness of clarification seems to be more dependent on the clarity of the original question.

\subsection{Performance on \texttt{Llama3-Test}}
We report the performance of the prompting-based and fine-tuned classifiers on \texttt{Llama3-Test} in Tables \ref{tab:llm_po_results} and \ref{tab:open_results}, respectively. In Table \ref{tab:llm_po_results}, the suffix ``-Gen'' indicates that the prompt optimization is performed on all of \texttt{Llama3-Train}, while the suffix ``-Lvl'' means that optimization only uses annotations for the specific pedagogical level. Note that SIMBA fails to identify a better-than-default prompt for level-specific \texttt{GPT-4o-mini} optimization, hence we do not include that result.


Both prompt-based and fine-tuned classifiers \textbf{perform decently at lower pedagogical levels}, but their \textbf{performance declines as the level increases}. Model differences are also more pronounced at higher levels: for example, \texttt{GPT-4o-mini} can outperform \texttt{GPT-4o} at lower levels, but \texttt{GPT-4o} excels at Levels 4 and 5. Similarly, fine-tuned Qwen models outperform frontier LLMs below Level 4 but not above.

\textbf{Prompt optimization can be effective for improving model performance on pedagogical assessment}, especially when level-specific annotations are used. We see that for Level 4, prompt optimization with level-specific training data leads to 28.4\% F-1 score improvement in \texttt{GPT-4o}. At higher pedagogical levels, prompt optimization consistently outperforms zero-shot LLM classifications, as a result of SIMBA adding beneficial few-shot examples.

\textbf{Fine-tuning on our synthetic data can boost model F-1 significantly}, by as much as 41.7\%. The fine-tuned ModernBERT models maintain competitive performance throughout most pedagogical levels compared to prompt-based classifiers. These results further validate the utility of synthetic data. Note that our data is synthesized with \texttt{GPT-4.1-nano}, which is not the most capable frontier model measured by MMLU\footnote{\url{https://openai.com/index/gpt-4-1/}}. It is possible that data generated from more powerful models can yield even more prominent performance gains. 

\subsection{Generalizability Analysis}

Given the consistent performance of our classifiers at Level 1, we focus our generalizability analysis on this level. One of the authors and one of the annotators annotated \texttt{New-LLM-Test} and \texttt{Llama3-MOOC} with random checks to ensure inter-rater reliability.

We report classifier performance on these datasets in Table \ref{tab:generalizability_results} and Table \ref{tab:generalizability_results_2}. A crucial difference between \texttt{Llama-3-70B-Instruct} and the other LLMs is the increased use of Markdown formatting in responses by the other LLMs. To assess whether such formatting influences classifier predictions, we remove Markdown formatting in out-of-distribution generations (\textbf{\ding{55} MD} columns).

Overall, there is a \textbf{decrease in classifier performance in both test datasets}, with over-scoring as a common error. Interestingly, stripping Markdown formatting can improve classifier performance on \texttt{New-LLM-Test} but reduce it on \texttt{Llama3-MOOC}, indicating \textbf{sensitivity to formatting cues}, which is a crucial consideration for classifier robustness. 

\textbf{Fine-tuned classifiers match the performance of LLMs on \texttt{Llama3-MOOC}}, showing robustness when tested on responses from the same model. In contrast, \textbf{prompt-based LLM classifiers are more generalizable across response sources}. Notably, \texttt{GPT-4o-Gen} performs best on out-of-distribution examples, implying that \textbf{prompt optimization across all pedagogical levels can enable \texttt{GPT-4o} to better model pedagogical reasoning}, resulting in improved generalization.


\begin{table}[!t]
    \centering
    \begin{tabular}{l|cccc}
    \toprule
    \textbf{Classifier} & \multicolumn{2}{c}{\textbf{Original}} & \multicolumn{2}{c}{\textbf{\ding{55} MD}}\\
     & F-1 & Acc. & F-1 & Acc. \\
    \midrule
    GPT-4o-mini & 63.1 & \textbf{73.3} & 64.0 & \textbf{73.3} \\
    GPT-4o-mini-Gen. & 64.1 & \textbf{73.3} & 63.8 & 72.8\\
    \midrule
    GPT-4o & 63.4 & 71.7 & 64.5 & 71.7\\
    GPT-4o-Gen. & \textbf{66.5} & 72.8 & \textbf{68.3} & \textbf{73.3} \\
    GPT-4o-Lvl. & 62.3 & 72.8 & 62.3 & 72.8\\
    \midrule
    FT-Qwen3-4B & 46.8 & 59.5 & 62.3 & 71.6 \\
    FT-Qwen3-8B & 48.4 & 61.0 & 64.0 & 72.8 \\
    FT-ModernBERT & 57.2 & 60.6 & 60.6 &  63.3\\
    \bottomrule
    \end{tabular}
    \caption{Performance of all of our classifiers on \texttt{New-LLM-Test} for Level 1. }
    \label{tab:generalizability_results}
\end{table}

\begin{table}[!t]
    \centering
    \begin{tabular}{l|cccc}
    \toprule
    \textbf{Classifier} & \multicolumn{2}{c}{\textbf{Original}} & \multicolumn{2}{c}{\textbf{\ding{55} MD}}\\
     & F-1 & Acc. & F-1 & Acc. \\
    \midrule
    GPT-4o-mini & 67.1 & 75.5 & 65.7 & 70.4 \\
    GPT-4o-mini-Gen. & 66.7 & 66.3 &69.5 & 69.4\\
    \midrule
    GPT-4o & 67.5 & 71.4 & 65.7 & 70.4 \\
    GPT-4o-Gen. &\textbf{72.8} & 76.1 & 70.1& 74.5 \\
    GPT-4o-Lvl. & 67.4 & 74.5 & 67.0 & 74.5 \\
    \midrule
    FT-Qwen3-4B & 72.2 & 76.7 & \textbf{70.8} & \textbf{76.7} \\
    FT-Qwen3-8B & 72.5 & \textbf{80.0} & 70.7 & \textbf{76.7} \\
    FT-ModernBERT & 68.4 & 74.5 & 63.4 & 65.3 \\
    \bottomrule
    \end{tabular}
    \caption{Performance of all of our classifiers on \texttt{Llama3-MOOC}.}
    \label{tab:generalizability_results_2}
\end{table}







\section{Discussion}
By analyzing VTA responses using our pedagogical framework and automating assessment with LLMs, we highlight key insights into the development and evaluation of VTAs in ALEs.

We inspect the consequences of incorporating peer forum posts to aid collaborative knowledge formation, under a simple VTA formulation with no course-specific materials involved. Our expert annotations suggest that including forum-level contexts can be detrimental to VTA performance in other pedagogical levels. This can be a result of the ``lost in the middle'' phenomenon \cite{liu2023lost}, where language models fixate on information presented at the beginning or the end of a long input. Since Level 5 appears at the end of our prompt, models may over-prioritize it. This underscores the value of our pedagogical framework, which makes such degradations in performance detectable. This result pertains only to \texttt{Llama-3-70B-Instruct}, although it is plausible that other LLMs would exhibit similar patterns.

Prior work \cite{maurya2024unifying} argues that LLMs may be unreliable annotators for pedagogical assessment. We explore this claim in the context of assessing VTAs in ALEs, and our findings largely support this concern. While prompt optimization and supervised fine-tuning on synthetic data successfully enhance accuracy, the models we examine still struggle with higher-level pedagogical evaluations. This could reflect a broader lack of pedagogical awareness during the training of state-of-the-art models \cite{macina2023mathdial, jurenka2024towards}, which impacts both prompt-based classification and data synthesis. 

Another limitation is that all our training data (and in-context examples for data generation) come from a single LLM, which might restrict generalization. Still, the fine-tuned classifiers match and outperform frontier LLM performance on \texttt{Llama3-MOOC}, suggesting strong potential. We expect that introducing more diverse VTA response styles to our training data will improve classifier robustness. We leave expanding expert annotations on a wider range of VTA outputs as future work. 

One interesting finding is that Markdown formatting from VTA responses affects classifier performance. We hypothesize that removing Markdown formatting brings the VTA responses closer to the output distribution of \texttt{Llama-3-70B-Instruct} for the open-weight classifiers, while mitigating self-selection biases in the prompt-based classifiers \cite{panickssery2024llm}. This is notable because it emphasizes a critical difference between human and model evaluators; humans are capable of disentangling stylistic features and pedagogical content, whereas current LLMs experience challenges.

\section{Conclusion}
We introduce the first evaluation framework for the pedagogical capabilities of teaching assistants in asynchronous online forums. Upon expert labeling of forum VTA responses using our rubric, we utilize these annotations to build and test automatic classifiers using language models. Our results indicate that while certain approaches can benefit classifier performance, language models continue to face challenges with higher-level, nuanced classifications and generalization across response types. Our novel framework paves the way for rigorous comparisons between different VTA systems, and we caution researchers when using LLMs for complex pedagogical evaluations.

\section*{Limitations}

Our pedagogical evaluation framework is grounded in educational theory, but its practical utility for instructors remains unclear. Future work will include usability studies with university instructors to better validate our framework's relevance and effectiveness in the real world.

Although the number of annotations is sufficient for prompt optimization in this work, a larger annotated dataset would always be desirable, as it can strengthen model training and evaluation. As an initial exploration into pedagogical assessment of VTAs, our work provides a solid foundation for future researchers in this domain. We encourage researchers to reuse our annotation framework to construct more training data for classifiers. The number of annotators was also limited; however, we achieved high inter-rater reliability. Additionally, recruiting expert annotators remains challenging in this domain, and we also do not provide financial compensation to our annotators due to budgetary concerns. Future work can look into expanding our annotator pool to enable annotating more examples.

This work focuses on simple VTA formulations that rely on models’ internal knowledge to answer academic questions, as we did not have access to vetted course materials. In contrast, existing VTAs often use retrieval-augmented generation (RAG) to obtain instructor-approved reference content. Studying such systems would offer valuable comparisons, but these implementations are typically not publicly available.

Lastly, our framework evaluates the pedagogical quality of VTA responses, not their factual accuracy or potential for hallucination. However, accurate and reliable content is critical in educational contexts. To be suitable for deployment, our pedagogical framework must be paired with methods that assess and promote response accuracy.

\section*{Ethical Considerations}
Asynchronous online forum posts made by students contain sensitive personal information such as their names and their grades (e.g. "I was wondering why I just got a B on this assignment"). Our suite of classifiers is designed to have access to the original student forum posts to enhance classification accuracy. While we redacted private information when developing our classifiers, it is important to recognize and minimize the potential risks of data leakage if our classifiers are adopted for real-world use. 
The most capable models we tested are closed-source and are only available through API access. Companies can train on the user data, and there can be attackers hacking into the data-hosting servers and gaining access to sensitive student data. In order to protect student privacy, future work should prioritize using open-source models securely hosted on local infrastructure.

\section*{Acknowledgment}
We would like to thank our annotators, Chenxi Shi and Xin Guan, for their insightful discussions and revision suggestions on the description of the evaluation framework, as well as for their careful and rigorous annotation work on this study.
\bibliography{custom}

\begin{thebibliography}{73}
\providecommand{\natexlab}[1]{#1}

\bibitem[{Agrawal et~al.(2015)Agrawal, Venkatraman, Leonard, and Paepcke}]{agrawal2015youedu}
Akshay Agrawal, Jagadish Venkatraman, Shane Leonard, and Andreas Paepcke. 2015.
\newblock Youedu: Addressing confusion in mooc discussion forums by recommending instructional video clips.

\bibitem[{Akcaoglu and Lee(2016)}]{akcaoglu2016increasing}
Mete Akcaoglu and Eunbae Lee. 2016.
\newblock Increasing social presence in online learning through small group discussions.
\newblock \emph{The international review of research in open and distributed learning}, 17(3).

\bibitem[{Akyol and Garrison(2011)}]{akyol2011assessing}
Zehra Akyol and D~Randy Garrison. 2011.
\newblock Assessing metacognition in an online community of inquiry.
\newblock \emph{The Internet and higher education}, 14(3):183--190.

\bibitem[{Al-Hossami et~al.(2024)Al-Hossami, Bunescu, Smith, and Teehan}]{al2024can}
Erfan Al-Hossami, Razvan Bunescu, Justin Smith, and Ryan Teehan. 2024.
\newblock Can language models employ the socratic method? experiments with code debugging.
\newblock In \emph{Proceedings of the 55th ACM Technical Symposium on Computer Science Education V. 1}, pages 53--59.

\bibitem[{Audras et~al.(2022)Audras, Zhao, Isgar, and Tang}]{audras2022virtual}
Diane Audras, Alex Zhao, Chuck Isgar, and Yucheng Tang. 2022.
\newblock Virtual teaching assistants: A survey of a novel teaching technology.
\newblock \emph{International Journal of Chinese Education}, 11(2):2212585X221121674.

\bibitem[{Baker et~al.(2024)Baker, Pankiewicz, and Liu}]{baker2024step}
Ryan~Shaun Baker, Maciej Pankiewicz, and Xiner Liu. 2024.
\newblock A step towards adaptive online learning: Exploring the role of gpt as virtual teaching assistants in online education.

\bibitem[{Brierton et~al.(2016)Brierton, Wilson, Kistler, Flowers, and Jones}]{brierton2016comparison}
Sara Brierton, Elizabeth Wilson, Mark Kistler, Jim Flowers, and David Jones. 2016.
\newblock A comparison of higher order thinking skills demonstrated in synchronous and asynchronous online college discussion posts.
\newblock \emph{Nacta Journal}, 60(1):14--21.

\bibitem[{Chan et~al.(2023)Chan, Chen, Su, Yu, Xue, Zhang, Fu, and Liu}]{chan2023chateval}
Chi-Min Chan, Weize Chen, Yusheng Su, Jianxuan Yu, Wei Xue, Shanghang Zhang, Jie Fu, and Zhiyuan Liu. 2023.
\newblock Chateval: Towards better llm-based evaluators through multi-agent debate.
\newblock \emph{arXiv preprint arXiv:2308.07201}.

\bibitem[{Chang et~al.(2024)Chang, Wang, Wang, Wu, Yang, Zhu, Chen, Yi, Wang, Wang et~al.}]{chang2024survey}
Yupeng Chang, Xu~Wang, Jindong Wang, Yuan Wu, Linyi Yang, Kaijie Zhu, Hao Chen, Xiaoyuan Yi, Cunxiang Wang, Yidong Wang, and 1 others. 2024.
\newblock A survey on evaluation of large language models.
\newblock \emph{ACM transactions on intelligent systems and technology}, 15(3):1--45.

\bibitem[{Chevalier et~al.(2024)Chevalier, Geng, Wettig, Chen, Mizera, Annala, Aragon, Fanlo, Frieder, Machado et~al.}]{chevalier2024language}
Alexis Chevalier, Jiayi Geng, Alexander Wettig, Howard Chen, Sebastian Mizera, Toni Annala, Max~Jameson Aragon, Arturo~Rodr{\'\i}guez Fanlo, Simon Frieder, Simon Machado, and 1 others. 2024.
\newblock Language models as science tutors.
\newblock \emph{arXiv preprint arXiv:2402.11111}.

\bibitem[{Chi and Wylie(2014)}]{chi2014icap}
Michelene~TH Chi and Ruth Wylie. 2014.
\newblock The icap framework: Linking cognitive engagement to active learning outcomes.
\newblock \emph{Educational psychologist}, 49(4):219--243.

\bibitem[{Choi et~al.(2005)Choi, Land, and Turgeon}]{choi2005scaffolding}
Ikseon Choi, Susan~M Land, and Alfred~J Turgeon. 2005.
\newblock Scaffolding peer-questioning strategies to facilitate metacognition during online small group discussion.
\newblock \emph{Instructional science}, 33(5):483--511.

\bibitem[{Dahal et~al.(2025)Dahal, Murray, Chataut, Hefeida, Srivastava, and Gyawali}]{dahal2025autota}
Rajashree Dahal, Greg Murray, Robin Chataut, Mohamed Hefeida, Anurag Srivastava, and Prashnna Gyawali. 2025.
\newblock Autota: A dynamic intent-based virtual teaching assistant for students using open source llms.
\newblock \emph{IEEE Access}.

\bibitem[{Dan et~al.(2023)Dan, Lei, Gu, Li, Yin, Lin, Ye, Tie, Zhou, Wang et~al.}]{dan2023educhat}
Yuhao Dan, Zhikai Lei, Yiyang Gu, Yong Li, Jianghao Yin, Jiaju Lin, Linhao Ye, Zhiyan Tie, Yougen Zhou, Yilei Wang, and 1 others. 2023.
\newblock Educhat: A large-scale language model-based chatbot system for intelligent education.
\newblock \emph{arXiv preprint arXiv:2308.02773}.

\bibitem[{Demszky et~al.(2021)Demszky, Liu, Mancenido, Cohen, Hill, Jurafsky, and Hashimoto}]{demszky2021measuring}
Dorottya Demszky, Jing Liu, Zid Mancenido, Julie Cohen, Heather Hill, Dan Jurafsky, and Tatsunori Hashimoto. 2021.
\newblock Measuring conversational uptake: A case study on student-teacher interactions.
\newblock \emph{arXiv preprint arXiv:2106.03873}.

\bibitem[{Dykman et~al.(2008)}]{dykman2008part}
Charlene~A Dykman and 1 others. 2008.
\newblock Part two-teaching online versus teaching conventionally.
\newblock \emph{Journal of Information Systems Education}, 19(2):157--164.

\bibitem[{Engle and Conant(2002)}]{engle2002guiding}
Randi~A Engle and Faith~R Conant. 2002.
\newblock Guiding principles for fostering productive disciplinary engagement: Explaining an emergent argument in a community of learners classroom.
\newblock \emph{Cognition and instruction}, 20(4):399--483.

\bibitem[{Farley and Burbules(2022)}]{farley2022online}
Irish~A Farley and Nicholas~C Burbules. 2022.
\newblock Online education viewed through an equity lens: Promoting engagement and success for all learners.
\newblock \emph{Review of Education}, 10(3):e3367.

\bibitem[{Fosnot(2013)}]{fosnot2013constructivism}
Catherine~Twomey Fosnot. 2013.
\newblock \emph{Constructivism: Theory, perspectives, and practice}.
\newblock Teachers College Press.

\bibitem[{Goel and Polepeddi(2018)}]{goel2018jill}
Ashok~K Goel and Lalith Polepeddi. 2018.
\newblock Jill watson: A virtual teaching assistant for online education.
\newblock In \emph{Learning engineering for online education}, pages 120--143. Routledge.

\bibitem[{Grattafiori et~al.(2024)Grattafiori, Dubey, Jauhri, Pandey, Kadian, Al-Dahle, Letman, Mathur, Schelten, Vaughan et~al.}]{grattafiori2024llama}
Aaron Grattafiori, Abhimanyu Dubey, Abhinav Jauhri, Abhinav Pandey, Abhishek Kadian, Ahmad Al-Dahle, Aiesha Letman, Akhil Mathur, Alan Schelten, Alex Vaughan, and 1 others. 2024.
\newblock The llama 3 herd of models.
\newblock \emph{arXiv preprint arXiv:2407.21783}.

\bibitem[{Han et~al.(2023)Han, Yoo, Myung, Kim, Lim, Kim, Lee, Hong, Kim, Ahn et~al.}]{han2023llm}
Jieun Han, Haneul Yoo, Junho Myung, Minsun Kim, Hyunseung Lim, Yoonsu Kim, Tak~Yeon Lee, Hwajung Hong, Juho Kim, So-Yeon Ahn, and 1 others. 2023.
\newblock Llm-as-a-tutor in efl writing education: Focusing on evaluation of student-llm interaction.
\newblock \emph{arXiv preprint arXiv:2310.05191}.

\bibitem[{Hew et~al.(2010)Hew, Cheung, and Ng}]{hew2010student}
Khe~Foon Hew, Wing~Sum Cheung, and Connie Siew~Ling Ng. 2010.
\newblock Student contribution in asynchronous online discussion: A review of the research and empirical exploration.
\newblock \emph{Instructional science}, 38:571--606.

\bibitem[{Hicke et~al.(2023)Hicke, Agarwal, Ma, and Denny}]{hicke2023ai}
Yann Hicke, Anmol Agarwal, Qianou Ma, and Paul Denny. 2023.
\newblock Ai-ta: Towards an intelligent question-answer teaching assistant using open-source llms.
\newblock \emph{arXiv preprint arXiv:2311.02775}.

\bibitem[{Joksimovi{\'c} et~al.(2015)Joksimovi{\'c}, Ga{\v{s}}evi{\'c}, Kovanovi{\'c}, Riecke, and Hatala}]{joksimovic2015social}
Sre{\'c}ko Joksimovi{\'c}, Dragan Ga{\v{s}}evi{\'c}, Vitomir Kovanovi{\'c}, Bernhard~E Riecke, and Marek Hatala. 2015.
\newblock Social presence in online discussions as a process predictor of academic performance.
\newblock \emph{Journal of Computer Assisted Learning}, 31(6):638--654.

\bibitem[{Jurenka et~al.(2024)Jurenka, Kunesch, McKee, Gillick, Zhu, Wiltberger, Phal, Hermann, Kasenberg, Bhoopchand et~al.}]{jurenka2024towards}
Irina Jurenka, Markus Kunesch, Kevin~R McKee, Daniel Gillick, Shaojian Zhu, Sara Wiltberger, Shubham~Milind Phal, Katherine Hermann, Daniel Kasenberg, Avishkar Bhoopchand, and 1 others. 2024.
\newblock Towards responsible development of generative ai for education: An evaluation-driven approach.
\newblock \emph{arXiv preprint arXiv:2407.12687}.

\bibitem[{Khattab et~al.(2023)Khattab, Singhvi, Maheshwari, Zhang, Santhanam, Vardhamanan, Haq, Sharma, Joshi, Moazam et~al.}]{khattab2023dspy}
Omar Khattab, Arnav Singhvi, Paridhi Maheshwari, Zhiyuan Zhang, Keshav Santhanam, Sri Vardhamanan, Saiful Haq, Ashutosh Sharma, Thomas~T Joshi, Hanna Moazam, and 1 others. 2023.
\newblock Dspy: Compiling declarative language model calls into self-improving pipelines.
\newblock \emph{arXiv preprint arXiv:2310.03714}.

\bibitem[{Kim and Kim(2024)}]{kim2024can}
Seungyoon Kim and Seungone Kim. 2024.
\newblock Can language models evaluate human written text? case study on korean student writing for education.
\newblock \emph{arXiv preprint arXiv:2407.17022}.

\bibitem[{Koo and Li(2016)}]{koo2016guideline}
Terry~K Koo and Mae~Y Li. 2016.
\newblock A guideline of selecting and reporting intraclass correlation coefficients for reliability research.
\newblock \emph{Journal of chiropractic medicine}, 15(2):155--163.

\bibitem[{Kweon et~al.(2025)Kweon, Nam, Lim, Hong, and Choi}]{kweon2025large}
Sunjun Kweon, Sooyohn Nam, Hyunseung Lim, Hwajung Hong, and Edward Choi. 2025.
\newblock A large-scale real-world evaluation of llm-based virtual teaching assistant.
\newblock \emph{arXiv preprint arXiv:2506.17363}.

\bibitem[{Laney and Dewan(2024)}]{laney2024human}
Mason Laney and Prasun Dewan. 2024.
\newblock Human-ai collaboration in a student discussion forum.
\newblock In \emph{Companion Proceedings of the 29th International Conference on Intelligent User Interfaces}, pages 74--77.

\bibitem[{Lantolf(2000)}]{lantolf2000introducing}
James~P Lantolf. 2000.
\newblock Introducing sociocultural theory.
\newblock \emph{Sociocultural theory and second language learning}, 1:1--26.

\bibitem[{Lee et~al.(2023)Lee, Myung, Han, Jin, and Oh}]{lee2023learning}
Changyoon Lee, Junho Myung, Jieun Han, Jiho Jin, and Alice Oh. 2023.
\newblock Learning from teaching assistants to program with subgoals: Exploring the potential for ai teaching assistants.
\newblock \emph{arXiv preprint arXiv:2309.10419}.

\bibitem[{Li et~al.(2023)Li, Qu, Shen, Min, and Yu}]{li2023curriculum}
Yu~Li, Shang Qu, Jili Shen, Shangchao Min, and Zhou Yu. 2023.
\newblock Curriculum-driven edubot: A framework for developing language learning chatbots through synthesizing conversational data.
\newblock \emph{arXiv preprint arXiv:2309.16804}.

\bibitem[{Lim{\'o}n(2001)}]{limon2001cognitive}
Margarita Lim{\'o}n. 2001.
\newblock On the cognitive conflict as an instructional strategy for conceptual change: A critical appraisal.
\newblock \emph{Learning and instruction}, 11(4-5):357--380.

\bibitem[{Lin and Gao(2020)}]{lin2020students}
Xi~Lin and Li~Gao. 2020.
\newblock Students’ sense of community and perspectives of taking synchronous and asynchronous online courses.
\newblock \emph{Asian Journal of Distance Education}, 15(1):169--179.

\bibitem[{Liu et~al.(2023)Liu, Lin, Hewitt, Paranjape, Bevilacqua, Petroni, and Liang}]{liu2023lost}
Nelson~F Liu, Kevin Lin, John Hewitt, Ashwin Paranjape, Michele Bevilacqua, Fabio Petroni, and Percy Liang. 2023.
\newblock Lost in the middle: How language models use long contexts.
\newblock \emph{arXiv preprint arXiv:2307.03172}.

\bibitem[{Liu et~al.(2024)Liu, Pankiewicz, Gupta, Huang, and Baker}]{liu2024step}
Xiner Liu, Maciej Pankiewicz, Tanvi Gupta, Zhongtian Huang, and Ryan~S Baker. 2024.
\newblock A step towards adaptive online learning: Exploring the role of gpt as virtual teaching assistants in online education.
\newblock \emph{Manuscript under review}.

\bibitem[{Liu et~al.(2025)Liu, Zambrano, Baker, Barany, Ocumpaugh, Zhang, Pankiewicz, Nasiar, and Wei}]{liu2025qualitative}
Xiner Liu, Andres~Felipe Zambrano, Ryan~S Baker, Amanda Barany, Jaclyn Ocumpaugh, Jiayi Zhang, Maciej Pankiewicz, Nidhi Nasiar, and Zhanlan Wei. 2025.
\newblock Qualitative coding with gpt-4: Where it works better.
\newblock \emph{Journal of Learning Analytics}, 12(1):169--185.

\bibitem[{Macina et~al.(2023)Macina, Daheim, Chowdhury, Sinha, Kapur, Gurevych, and Sachan}]{macina2023mathdial}
Jakub Macina, Nico Daheim, Sankalan~Pal Chowdhury, Tanmay Sinha, Manu Kapur, Iryna Gurevych, and Mrinmaya Sachan. 2023.
\newblock Mathdial: A dialogue tutoring dataset with rich pedagogical properties grounded in math reasoning problems.
\newblock \emph{arXiv preprint arXiv:2305.14536}.

\bibitem[{Maurya et~al.(2024)Maurya, Srivatsa, Petukhova, and Kochmar}]{maurya2024unifying}
Kaushal~Kumar Maurya, KV~Srivatsa, Kseniia Petukhova, and Ekaterina Kochmar. 2024.
\newblock Unifying ai tutor evaluation: An evaluation taxonomy for pedagogical ability assessment of llm-powered ai tutors.
\newblock \emph{arXiv preprint arXiv:2412.09416}.

\bibitem[{McLoughlin and Mynard(2009)}]{mcloughlin2009analysis}
David McLoughlin and Jo~Mynard. 2009.
\newblock An analysis of higher order thinking in online discussions.
\newblock \emph{Innovations in Education and Teaching International}, 46(2):147--160.

\bibitem[{Meyer(2003)}]{meyer2003face}
Katrina~A Meyer. 2003.
\newblock Face-to-face versus threaded discussions: The role of time and higher-order thinking.
\newblock \emph{Journal of asynchronous learning networks}, 7(3):55--65.

\bibitem[{Michalsky et~al.(2007)Michalsky, Zion, and Mevarech}]{michalsky2007developing}
Tova Michalsky, Michal Zion, and Zemira~R Mevarech. 2007.
\newblock Developing students' metacognitive awareness in asynchronous learning networks in comparison to face-to-face discussion groups.
\newblock \emph{Journal of Educational Computing Research}, 36(4):395--424.

\bibitem[{Murphy(2009)}]{murphy2009framework}
Elizabeth Murphy. 2009.
\newblock A framework for identifying and promoting metacognitive knowledge and control in online discussants.
\newblock \emph{Canadian Journal of Learning and Technology/La revue canadienne de l’apprentissage et de la technologie}, 34(2).

\bibitem[{Nandi et~al.(2012)Nandi, Hamilton, and Harland}]{nandi2012evaluating}
Dip Nandi, Margaret Hamilton, and James Harland. 2012.
\newblock Evaluating the quality of interaction in asynchronous discussion forums in fully online courses.
\newblock \emph{Distance education}, 33(1):5--30.

\bibitem[{Onyema et~al.(2019)Onyema, Deborah, Alsayed, Noorulhasan, and Sanober}]{onyema2019online}
Edeh~Michael Onyema, Edeh~Chinecherem Deborah, Alhuseen~Omar Alsayed, Quadri Noorulhasan, and Sumaya Sanober. 2019.
\newblock Online discussion forum as a tool for interactive learning and communication.
\newblock \emph{International Journal of Recent Technology and Engineering}, 8(4):4852--4859.

\bibitem[{Panickssery et~al.(2024)Panickssery, Bowman, and Feng}]{panickssery2024llm}
Arjun Panickssery, Samuel Bowman, and Shi Feng. 2024.
\newblock Llm evaluators recognize and favor their own generations.
\newblock \emph{Advances in Neural Information Processing Systems}, 37:68772--68802.

\bibitem[{Pena-Shaff and Nicholls(2004)}]{pena2004analyzing}
Judith~B Pena-Shaff and Craig Nicholls. 2004.
\newblock Analyzing student interactions and meaning construction in computer bulletin board discussions.
\newblock \emph{Computers \& Education}, 42(3):243--265.

\bibitem[{Qiao et~al.(2025)Qiao, Denny, and Giacaman}]{qiao2025oversight}
Shuying Qiao, Paul Denny, and Nasser Giacaman. 2025.
\newblock Oversight in action: Experiences with instructor-moderated llm responses in an online discussion forum.
\newblock In \emph{Proceedings of the 27th Australasian Computing Education Conference}, pages 95--104.

\bibitem[{Raamadhurai et~al.(2019)Raamadhurai, Baker, and Poduval}]{raamadhurai2019curio}
Srikrishna Raamadhurai, Ryan Baker, and Vikraman Poduval. 2019.
\newblock Curio smartchat: a system for natural language question answering for self-paced k-12 learning.
\newblock In \emph{Proceedings of the Fourteenth Workshop on Innovative Use of NLP for Building Educational Applications}, pages 336--342.

\bibitem[{Sajja et~al.(2023)Sajja, Sermet, Cwiertny, and Demir}]{sajja2023platform}
Ramteja Sajja, Yusuf Sermet, David Cwiertny, and Ibrahim Demir. 2023.
\newblock Platform-independent and curriculum-oriented intelligent assistant for higher education.
\newblock \emph{International journal of educational technology in higher education}, 20(1):42.

\bibitem[{Sinha et~al.(2024)Sinha, Goyal, Sy, Kuperus, Dickey, and Bejarano}]{sinha2024boilertai}
Anvit Sinha, Shruti Goyal, Zachary Sy, Rhianna Kuperus, Ethan Dickey, and Andres Bejarano. 2024.
\newblock Boilertai: A platform for enhancing instruction using generative ai in educational forums.
\newblock In \emph{2024 IEEE Frontiers in Education Conference (FIE)}, pages 1--8. IEEE.

\bibitem[{Stahl(2013)}]{stahl2013model}
Gerry Stahl. 2013.
\newblock A model of collaborative knowledge-building.
\newblock In \emph{International conference of the learning sciences}, pages 70--77. Psychology Press.

\bibitem[{Szabo and Schwartz(2011)}]{szabo2011learning}
Zsuzsanna Szabo and Jonathan Schwartz. 2011.
\newblock Learning methods for teacher education: The use of online discussions to improve critical thinking.
\newblock \emph{Technology, Pedagogy and Education}, 20(1):79--94.

\bibitem[{Tack and Piech(2022)}]{tack2022ai}
Ana{\"\i}s Tack and Chris Piech. 2022.
\newblock The ai teacher test: Measuring the pedagogical ability of blender and gpt-3 in educational dialogues.
\newblock \emph{arXiv preprint arXiv:2205.07540}.

\bibitem[{Taneja et~al.(2024)Taneja, Maiti, Kakar, Guruprasad, Rao, and Goel}]{taneja2024jill}
Karan Taneja, Pratyusha Maiti, Sandeep Kakar, Pranav Guruprasad, Sanjeev Rao, and Ashok~K Goel. 2024.
\newblock Jill watson: A virtual teaching assistant powered by chatgpt.
\newblock In \emph{International Conference on Artificial Intelligence in Education}, pages 324--337. Springer.

\bibitem[{Tate and Warschauer(2022)}]{tate2022equity}
Tamara Tate and Mark Warschauer. 2022.
\newblock Equity in online learning.
\newblock \emph{Educational Psychologist}, 57(3):192--206.

\bibitem[{Vasselli et~al.(2023)Vasselli, Vasselli, Nohejl, and Watanabe}]{vasselli2023naisteacher}
Justin Vasselli, Christopher Vasselli, Adam Nohejl, and Taro Watanabe. 2023.
\newblock Naisteacher: A prompt and rerank approach to generating teacher utterances in educational dialogues.
\newblock In \emph{Proceedings of the 18th Workshop on Innovative Use of NLP for Building Educational Applications (BEA 2023)}, pages 772--784.

\bibitem[{Wang et~al.(2023{\natexlab{a}})Wang, Li, Chen, Cai, Zhu, Lin, Cao, Liu, Liu, and Sui}]{wang2023large}
Peiyi Wang, Lei Li, Liang Chen, Zefan Cai, Dawei Zhu, Binghuai Lin, Yunbo Cao, Qi~Liu, Tianyu Liu, and Zhifang Sui. 2023{\natexlab{a}}.
\newblock Large language models are not fair evaluators.
\newblock \emph{arXiv preprint arXiv:2305.17926}.

\bibitem[{Wang et~al.(2023{\natexlab{b}})Wang, Zhang, Robinson, Loeb, and Demszky}]{wang2023bridging}
Rose~E Wang, Qingyang Zhang, Carly Robinson, Susanna Loeb, and Dorottya Demszky. 2023{\natexlab{b}}.
\newblock Bridging the novice-expert gap via models of decision-making: A case study on remediating math mistakes.
\newblock \emph{arXiv preprint arXiv:2310.10648}.

\bibitem[{Wang et~al.(2016)Wang, Wen, and Ros{\'e}}]{wang2016towards}
Xu~Wang, Miaomiao Wen, and Carolyn~P Ros{\'e}. 2016.
\newblock Towards triggering higher-order thinking behaviors in moocs.
\newblock In \emph{Proceedings of the sixth international conference on learning analytics \& knowledge}, pages 398--407.

\bibitem[{Warner et~al.(2024)Warner, Chaffin, Clavi{\'e}, Weller, Hallstr{\"o}m, Taghadouini, Gallagher, Biswas, Ladhak, Aarsen et~al.}]{warner2024smarter}
Benjamin Warner, Antoine Chaffin, Benjamin Clavi{\'e}, Orion Weller, Oskar Hallstr{\"o}m, Said Taghadouini, Alexis Gallagher, Raja Biswas, Faisal Ladhak, Tom Aarsen, and 1 others. 2024.
\newblock Smarter, better, faster, longer: A modern bidirectional encoder for fast, memory efficient, and long context finetuning and inference.
\newblock \emph{arXiv preprint arXiv:2412.13663}.

\bibitem[{Watts(2016)}]{watts2016synchronous}
Lynette Watts. 2016.
\newblock Synchronous and asynchronous communication in distance learning: A review of the literature.
\newblock \emph{Quarterly Review of Distance Education}, 17(1):23.

\bibitem[{Wei et~al.(2025)Wei, Wang, Bi, Chen, Li, Jiang, Lin, Zhang, Song, Li et~al.}]{wei2025elmes}
Shou'ang Wei, Xinyun Wang, Shuzhen Bi, Jian Chen, Ruijia Li, Bo~Jiang, Xin Lin, Min Zhang, Yu~Song, BingDong Li, and 1 others. 2025.
\newblock Elmes: An automated framework for evaluating large language models in educational scenarios.
\newblock \emph{arXiv preprint arXiv:2507.22947}.

\bibitem[{Xu et~al.(2025)Xu, Li, Huan, Minaya, and Yu}]{xu2025course}
Zhen Xu, Xinjin Li, Yingqi Huan, Veronica Minaya, and Renzhe Yu. 2025.
\newblock From course to skill: Evaluating large language model performance in curricular analytics.
\newblock In \emph{International Conference on Artificial Intelligence in Education}, pages 203--211. Springer.

\bibitem[{Yang et~al.(2025)Yang, Li, Yang, Zhang, Hui, Zheng, Yu, Gao, Huang, Lv et~al.}]{yang2025qwen3}
An~Yang, Anfeng Li, Baosong Yang, Beichen Zhang, Binyuan Hui, Bo~Zheng, Bowen Yu, Chang Gao, Chengen Huang, Chenxu Lv, and 1 others. 2025.
\newblock Qwen3 technical report.
\newblock \emph{arXiv preprint arXiv:2505.09388}.

\bibitem[{Yang et~al.(2005)Yang, Newby, and Bill}]{yang2005using}
Ya-Ting~C Yang, Timothy~J Newby, and Robert~L Bill. 2005.
\newblock Using socratic questioning to promote critical thinking skills through asynchronous discussion forums in distance learning environments.
\newblock \emph{The american journal of distance education}, 19(3):163--181.

\bibitem[{Yang et~al.(2022)Yang, Yuan, Feng, Li, and van Aalst}]{yang2022fostering}
Yuqin Yang, Kaicheng Yuan, Xueqi Feng, Xiuhan Li, and Jan van Aalst. 2022.
\newblock Fostering low-achieving students' productive disciplinary engagement through knowledge-building inquiry and reflective assessment.
\newblock \emph{British Journal of Educational Technology}, 53(6):1511--1529.

\bibitem[{Zamfirescu-Pereira et~al.(2025)Zamfirescu-Pereira, Qi, Hartmann, DeNero, and Norouzi}]{zamfirescu202561a}
JD~Zamfirescu-Pereira, Laryn Qi, Bj{\"o}rn Hartmann, John DeNero, and Narges Norouzi. 2025.
\newblock 61a bot report: Ai assistants in cs1 save students homework time and reduce demands on staff.(now what?).
\newblock In \emph{Proceedings of the 56th ACM Technical Symposium on Computer Science Education V. 1}, pages 1309--1315.

\bibitem[{Zhang et~al.(2024{\natexlab{a}})Zhang, Chen, and Yu}]{zhang2024prolex}
Xuanming Zhang, Zixun Chen, and Zhou Yu. 2024{\natexlab{a}}.
\newblock Prolex: A benchmark for language proficiency-oriented lexical substitution.
\newblock \emph{arXiv preprint arXiv:2401.11356}.

\bibitem[{Zhang et~al.(2024{\natexlab{b}})Zhang, Diaz, Chen, Wu, Qian, Voss, and Yu}]{zhang2024decor}
Xuanming Zhang, Anthony Diaz, Zixun Chen, Qingyang Wu, Kun Qian, Erik Voss, and Zhou Yu. 2024{\natexlab{b}}.
\newblock Decor: Improving coherence in l2 english writing with a novel benchmark for incoherence detection, reasoning, and rewriting.
\newblock \emph{arXiv preprint arXiv:2406.19650}.

\bibitem[{Zou et~al.(2021)Zou, Hu, Pan, Li, Cai, and Liu}]{zou2021exploring}
Wenting Zou, Xiao Hu, Zilong Pan, Chenglu Li, Ying Cai, and Min Liu. 2021.
\newblock Exploring the relationship between social presence and learners’ prestige in mooc discussion forums using automated content analysis and social network analysis.
\newblock \emph{Computers in Human Behavior}, 115:106582.

\end{thebibliography}

\appendix

\section{Use of AI Assistants}

We used AI Assistants to help polish sections of the paper after completing an initial draft of the paper.

\section{Scoring Rubrics}
\label{sec:rubrics}

Please refer to Table \ref{tab:rubric} for the full rubric for our pedagogical framework and Table \ref{tab:example1}, Table \ref{tab:example2}, and Table \ref{tab:example3} for scoring examples. All examples are rewritten based on the human annotation data for privacy consideration.

\section{Virtual Teaching Assistant Prompts}
\label{sec:vta_prompts}

\lstset{
basicstyle=\small\ttfamily,
columns=flexible,
breaklines=true
}

\subsection{VTA Prompt without Context}
\begin{lstlisting}
"""You are a virtual teaching assistant for a course called <COURSE_NAME>. 
<COURSE_DESCRIPTION>
Respond to the discussion forum post for this course provided by one of the students. Please offer the response based on your existing knowledge base. Please add a general greeting in each response.

Please adhere to the following pedagogical goals:
1. Clarify Misunderstandings: Support knowledge acquisition by articulating questions, addressing confusion, and receiving clarifications from peers or instructors.
2. Deepen Disciplinary Understanding: Promote deeper engagement with core concepts and themes through elaboration, critical questioning, and interaction with diverse perspectives.
3. Develop Higher-Order Thinking: Cultivate critical thinking and reasoning skills by analyzing ideas, justifying positions, synthesizing information, and exploring alternative viewpoints.
4. Enhance Metacognitive Awareness: Strengthen self-regulated learning by reflecting on one's understanding, identifying gaps in knowledge, and evaluating the quality of reasoning."""
\end{lstlisting}

\subsection{VTA Prompt with Forum-level Context}

\begin{lstlisting}
    
"""You are a virtual teaching assistant for a course called <COURSE_NAME>. 
<COURSE_DESCRIPTION>

You are responding to the following discussion forum post for this course provided by one of the students. This post is an initial post in the discussion.

Additionally, here are some other relevant posts from students on this same topic:
<SIMILAR_POSTS>

Please offer the response based on your existing knowledge base. Please add a general greeting in each response.

When appropriate, refer to insights or perspectives from these related posts to foster connections between student ideas.

Please adhere to the following pedagogical goals:
1. Clarify Misunderstandings: Support knowledge acquisition by articulating questions, addressing confusion, and receiving clarifications from peers or instructors.
2. Deepen Disciplinary Understanding: Promote deeper engagement with core concepts and themes through elaboration, critical questioning, and interaction with diverse perspectives.
3. Develop Higher-Order Thinking: Cultivate critical thinking and reasoning skills by analyzing ideas, justifying positions, synthesizing information, and exploring alternative viewpoints.
4. Enhance Metacognitive Awareness: Strengthen self-regulated learning by reflecting on one's understanding, identifying gaps in knowledge, and evaluating the quality of reasoning.
5. Foster Collaborative Knowledge Construction and Social Presence: Fosters peer interaction and collective learning by connecting diverse student perspectives, encouraging the exchange of ideas, and supporting collaborative knowledge construction, positioning the discussion forum as a shared space for dialogue and co-construction of understanding."""

\end{lstlisting}

\subsection{VTA Prompt for \texttt{Llama3-MOOC}}

\begin{lstlisting}
"""You are a virtual teaching assistant. Respond to the discussion forum post provided by one of the students. Please offer the response based on your existing knowledge base. Please add a general greeting in each response.

Please adhere to the following pedagogical goals:
1. Clarify Misunderstandings: Support knowledge acquisition by articulating questions, addressing confusion, and receiving clarifications from peers or instructors.
2. Deepen Disciplinary Understanding: Promote deeper engagement with core concepts and themes through elaboration, critical questioning, and interaction with diverse perspectives.
3. Develop Higher-Order Thinking: Cultivate critical thinking and reasoning skills by analyzing ideas, justifying positions, synthesizing information, and exploring alternative viewpoints.
4. Enhance Metacognitive Awareness: Strengthen self-regulated learning by reflecting on one's understanding, identifying gaps in knowledge, and evaluating the quality of reasoning."""

\end{lstlisting}

\section{Annotator Demographic Information}
Both annotators are Asian females with an average age of 23. They are both masters students in the field of education enrolled in Teachers College.

\section{Synthetic Data Details}
\label{sec:synth_data}

Our detailed process for generating synthetic data is as follows:

\begin{enumerate}
    \item \textit{Extract the description for each level-rating combination from the standardized rubric.} For example, the description for Level 3's ``Not Present (0)'' would be ``No effort is made to promote higher-order thinking.''
    \item \textit{Repeatedly sample in-context examples for data synthesis.} To avoid data contamination, we sample exclusively from annotations in \texttt{Llama3-Train}. For every call to \texttt{GPT-4.1-nano}, we use five examples to produce three new post-response pairs. 
    
    Furthermore, we prioritize sampling from the same level-rating combination if possible to promote synthetic data quality: if the number of post-response pairs in a specific combination exceeds our generation needs, we restrict sampling to that subset; otherwise, we sample from all post-response pairs.
    \item \textit{Synthesize new post-response pairs from in-context examples and VTA response requirements.} We include the description from the first step as the requirement for the VTA response in the newly synthesized post-response pair. For every level-rating combination, we synthesize up to 300 instances; if this level-rating combination already has more than 300 instances, no additional examples are created. 
\end{enumerate}

\section{Model Training Details}
\label{sec: model_training}
All models are trained for 10 epochs on an A100 GPU (80 GB GPU memory) with an initial learning rate of $2e\text{-}5$, and the best checkpoints are selected using validation loss. We set the maximum token limit for ModernBERT to be 8192. 

See the prompts used to perform supervised fine-tuning for Qwen models in Appendix \ref{sec:sft_prompt_qwen}.

\newpage
\onecolumn

\section{Prompt for Supervised Fine-tuning of Qwen3-4B and Qwen3-8B}
\label{sec:sft_prompt_qwen}
\begin{lstlisting}
Given the following discussion forum post from a student and the response from a teaching assistant:

---
<POST_RESPONSE_PAIR>
---

Assess the response from the teaching assistant with the following rubric:

---
<RUBRIC>
---

Provide your rating directly as "0", "1", "2", or "NA".
\end{lstlisting}

\section{Summary Statistics of the Annotation Data}

Refer to Tables \ref{tab:score with context} and \ref{tab:score without context} for detailed statistics on \texttt{Llama-3-70B-Instruct} annotations.

\begin{table}[!ht]
\centering
\small
\setlength{\tabcolsep}{5pt}
\caption{Annotation score distribution without context inclusion}
    \begin{tabular}{lcccc}
    \toprule
         &  NA&  0&  1&  2\\
         \midrule
         Clarify misun
derstandings&  0&  0&  19&  281\\
         Disciplinary
Understanding &  2&  1&  50&  247\\
         Higher-Order
Thinking&  3&  22&  131&  144\\
         Metacognitive
Awareness&  2&  27&  222&  49\\

\bottomrule
\label{tab:score without context}
\end{tabular}
\end{table}

\begin{table}[!ht]
\centering
\small
\setlength{\tabcolsep}{5pt}
\caption{Annotation score distribution with context inclusion}
    \begin{tabular}{lcccc}
    \toprule
         &  NA&  0&  1&  2\\
         \midrule
         Clarify misun
derstandings&  0&  0&  67&  221\\
         Disciplinary
Understanding &  0&  29&  175&  93\\
         Higher-Order
Thinking&  0&  58&  137&  100\\
         Metacognitive
Awareness&  0&  108&  140&  50\\
Collaborative Knowledge \\ Construction & 0& 42& 167& 89\\
\bottomrule
\end{tabular}
\label{tab:score with context}
\end{table}

\section{Example Generations}
\label{sec:example_gens}

We present example generations of different LLMs under context-free and forum-level context conditions for the same student forum post. To protect student privacy, we omit the original student post here. We additionally redact any names mentioned in the simulated VTA responses.

\subsection{Llama-3-70B-Instruct, Context-Free}

\begin{lstlisting}
Hello everyone,

Thank you for sharing your thoughts and questions about Chapter 14, particularly regarding bipolar disorder. The statistic you mentioned is indeed alarming, and it's essential to understand the underlying factors that contribute to the increased risk of suicide in individuals with bipolar disorder.

Research suggests that the transition from a manic episode to a depressive episode, often referred to as a "crash" or "comedown," can be a particularly vulnerable time for individuals with bipolar disorder. Several factors may contribute to this increased risk:

1. **Mood instability**: The rapid shift from a manic to a depressive state can be overwhelming, leading to feelings of hopelessness, despair, and anxiety.
2. **Loss of control**: During a manic episode, individuals may feel invincible or in control. However, as the episode subsides, they may experience a sense of loss of control, leading to feelings of powerlessness and helplessness.
3. **Cognitive distortions**: Negative thought patterns, such as rumination and self-blame, can become more pronounced during the depressive phase, increasing the risk of suicidal ideation.
4. **Neurobiological factors**: The crash from a manic episode can be accompanied by changes in neurotransmitter levels, such as a decrease in dopamine and serotonin, which can contribute to feelings of depression and hopelessness.
5. **Lack of coping skills**: Individuals with bipolar disorder may struggle with developing effective coping strategies to manage the transition from a manic to a depressive episode, increasing their vulnerability to suicidal thoughts.

It's essential to note that these factors can vary from person to person, and not everyone with bipolar disorder will experience suicidal ideation. However, it's crucial to acknowledge the increased risk and provide support and resources to individuals struggling with the disorder.

If anyone with bipolar disorder or experience with the condition would like to share their insights, I would be grateful to hear their perspectives. Additionally, I encourage everyone to consider the importance of seeking help and support for individuals struggling with mental health conditions.

Best regards,
\end{lstlisting}

\subsection{Llama-3-70B-Instruct, Forum-Level Context}
\begin{lstlisting}
Hello [REDACTED],

Thank you for sharing your question about bipolar disorder. It's indeed a striking statistic that 20% of people with the illness die from suicide. Your question about what brings a person coming out of a manic episode to that level is a crucial one.

From what we've learned in Chapter 14, it's possible that the crash or comedown from a manic episode can lead to feelings of intense despair, hopelessness, and anxiety. This can be a particularly vulnerable time for individuals with bipolar disorder, as they may feel like they're losing control or experiencing a sense of emotional numbness.

[REDACTED]'s response in Similar Post #2 provides some insight into the intrusive nature of manic episodes, which can lead to self-destructive behavior. It's also important to note that individuals with bipolar disorder often experience a mix of symptoms, including depression, anxiety, and impulsivity, which can contribute to suicidal ideation.

It would be great to hear from others who may have personal experience with bipolar disorder or have insights into this topic. Additionally, it might be helpful to explore the role of therapy, support systems, and medication in managing symptoms and reducing the risk of suicide.

What do you think about the importance of early intervention and treatment in preventing suicidal ideation in individuals with bipolar disorder?

Best, [Your Name]
\end{lstlisting}

\subsection{GPT-4.1, Context-Free}
\begin{lstlisting}
Hello! Thank you for your thoughtful post and for raising such an important question about bipolar disorder.

You're absolutely right - bipolar disorder carries a significant risk for suicide, and it's a topic that deserves careful attention and understanding. Your observation about suicides often occurring during the ``comedown'' from a manic episode is supported by research. Let's break down why this might happen:

**1. The Emotional Crash:**  
After a manic episode, individuals often experience a sharp drop in mood, sometimes into a depressive episode. This transition can feel especially devastating because the contrast between the high energy, optimism, and grandiosity of mania and the hopelessness of depression is so stark. The sudden loss of energy and motivation, combined with feelings of guilt or regret about actions taken during mania, can be overwhelming.

**2. Increased Insight:**  
During mania, people may not fully recognize the consequences of their actions. As they come down, they might gain insight into the problems caused during the manic phase (e.g., financial issues, damaged relationships), which can lead to intense feelings of shame, guilt, or hopelessness.

**3. Energy Levels:**  
Interestingly, people in the depths of depression may lack the energy to act on suicidal thoughts. However, during the transition out of mania, energy levels may still be relatively high, but mood and outlook have plummeted. This combination can increase the risk of acting on suicidal impulses.

**4. Biological Factors:**  
There are also neurobiological changes happening in the brain during these transitions, which can affect mood regulation and impulse control.

Your openness to hearing from those with lived experience is wonderful, as personal perspectives can add depth to our understanding. If anyone in the class feels comfortable sharing, their insights could be very valuable.

**Reflective Question:**  
How do you think understanding these patterns might influence the way clinicians or loved ones support someone with bipolar disorder during these transitions? What kinds of interventions or supports do you think could help reduce risk during these vulnerable periods?

Let me know if you'd like more information on this topic or if you have other questions about Chapter 14!
\end{lstlisting}

\subsection{GPT-4.1, Forum-Level Context}
\begin{lstlisting}
Hello! Thank you for sharing your thoughts and raising such an important and sensitive question about bipolar disorder. Your observation about the high risk of suicide, especially during the transition out of a manic episode, is both accurate and deeply concerning.

To help clarify, when someone with bipolar disorder is coming down from a manic episode (sometimes called the "post-manic crash"), they often experience a rapid and severe shift in mood. During mania, individuals may feel euphoric, energetic, and invincible, sometimes making risky decisions or acting impulsively. When the mania ends, the contrast can be stark - they may suddenly feel exhausted, ashamed, or overwhelmed by the consequences of their actions during the manic phase. This abrupt change can lead to intense feelings of hopelessness, guilt, or depression, which can increase the risk of suicidal thoughts or behaviors.

Your question also connects with some of the insights from other students in this discussion. For example, [REDACTED] (Similar post #2) mentioned how intrusive mania can be and how the phases of bipolar disorder differ from other disorders like borderline personality disorder. This highlights the unique challenges of mood cycling in bipolar disorder, which can make the emotional lows feel even more severe after a high.

It's also important to note that during the "comedown," individuals may regain enough energy and clarity to act on suicidal thoughts that may have been present during the depressive phase but were not acted upon due to lack of motivation or energy. This is one reason why the period following mania is particularly risky.

If you or anyone you know is struggling with these issues, it's crucial to seek support from mental health professionals. Understanding these patterns can help us be more compassionate and supportive toward those living with bipolar disorder.

Does this help clarify your question? Are there any other aspects of bipolar disorder or mood disorders you'd like to explore further? Your curiosity and willingness to engage with these difficult topics are really valuable for our collective learning.
\end{lstlisting}

\subsection{GPT-4o, Context-Free}

\begin{lstlisting}
Hello!

It's great that you're engaging deeply with the material on bipolar disorder. The statistic you mentioned highlights the serious nature of the disorder and the importance of understanding its complexities.

To address your question, individuals with bipolar disorder can experience extreme mood swings, including manic, hypomanic, and depressive episodes. During a manic episode, a person might feel euphoric, have increased energy, and engage in risky behaviors. However, when transitioning out of a manic episode, individuals often experience a "crash" or a significant drop in mood, which can lead to feelings of depression, hopelessness, and despair.

This transition can be particularly challenging because the contrast between the high energy and optimism of mania and the low energy and negativity of depression can be stark and overwhelming. Additionally, during the comedown, individuals might reflect on the consequences of their actions during the manic phase, which can exacerbate feelings of guilt or shame.

It's important to note that each person's experience with bipolar disorder is unique, and various factors, including biological, psychological, and social influences, can contribute to these experiences. If anyone with personal experience wishes to share their insights, it could provide valuable perspectives.

If you have further questions or need clarification, feel free to ask. Understanding these nuances can help in developing empathy and effective support strategies for those affected by bipolar disorder.

Best regards!
\end{lstlisting}

\subsection{GPT-4o, Forum-Level Context}

\begin{lstlisting}
Hello!

Your question about the transition from a manic episode to a depressive state in individuals with bipolar disorder is an important one. This transition can be particularly challenging and is often associated with a heightened risk of suicide. During a manic episode, individuals may experience elevated mood, increased energy, and a sense of invincibility. However, as they come down from this state, they may face a stark contrast in mood, leading to feelings of hopelessness, guilt, or worthlessness. This sudden shift can be overwhelming and may contribute to suicidal thoughts or behaviors.

It's also important to consider that during the comedown, individuals might reflect on the consequences of their actions during the manic phase, which can lead to regret or distress. Additionally, the biochemical changes in the brain during these transitions can exacerbate mood instability.

In the related posts, [REDACTED] mentions the intrusive nature of mania in bipolar I disorder, which can lead to self-destructive behavior. This aligns with the idea that the aftermath of such behavior can be difficult to cope with, potentially leading to suicidal ideation.

Understanding these dynamics is crucial for providing support and intervention. If you or someone you know is struggling, it's important to seek professional help. Mental health professionals can offer strategies and treatments to manage these transitions more effectively.

Feel free to continue the discussion or ask more questions if you have them!

Best regards.
\end{lstlisting}

\subsection{Qwen3-32B, Context-Free}
We remove the thinking process of \texttt{Qwen3-32B}.

\begin{lstlisting}
**Hello!** Thank you for sharing such a thoughtful and important question. You're absolutely right to highlight the connection between bipolar disorder and suicide risk, particularly during the transition from mania to depression. Let's unpack this step by step:  

1. **The "Comedown" Phase and Emotional Vulnerability**:  
   During a manic episode, individuals often feel invincible, hyper-focused, or euphoric, which can mask their awareness of distress. However, the crash after mania (often a depressive episode) can bring intense emotional pain. This phase may involve guilt, shame, or despair over the consequences of impulsive actions during mania (e.g., financial ruin, strained relationships). These feelings can feel insurmountable, increasing suicide risk.  

2. **Biological and Neurochemical Factors**:  
   The transition between mood states in bipolar disorder is linked to dysregulation in neurotransmitters like serotonin and dopamine. This instability can amplify emotional reactivity and hopelessness during the depressive phase. Additionally, sleep deprivation during mania (a common symptom) can exacerbate irritability and impulsivity during the comedown.  

3. **Mixed Episodes**:  
   Some individuals experience *mixed episodes*, where manic and depressive symptoms co-occur. This combination - e.g., feeling agitated, restless, and hopeless simultaneously - can heighten suicide risk, as the person may lack the energy to seek help but feel overwhelmed by their suffering.  

4. **Stigma and Treatment Gaps**:  
   People with bipolar disorder may not seek help during mania (due to denial of illness) or during the depressive phase (due to low motivation or stigma). Without consistent treatment (e.g., mood stabilizers, therapy), the cycle of mania-depression can worsen, increasing risk.  

**Critical Question for Reflection**: How might societal misconceptions about bipolar disorder (e.g., ``they're just moody'' or ``mania is fun'') contribute to underdiagnosis or undertreatment during critical phases?  

If you or someone you know is struggling with these issues, please emphasize the importance of professional support. For further discussion, I'd be happy to explore how therapy (e.g., CBT, DBT) or medication management might mitigate these risks. Let me know if you'd like to dive deeper!  

- Your Teaching Assistant
\end{lstlisting}

\subsection{Qwen3-32B, Forum-Level Context}

\begin{lstlisting}
**Hello [REDACTED],**  
Thank you for raising such an important and thought-provoking question about bipolar disorder and suicide risk. Your observation about the 20% suicide rate is striking, and your focus on the ``comedown'' from mania is a critical point. Let's unpack this together.  

### **Understanding the Transition Phase**  
During a manic episode, individuals often experience heightened energy, impulsivity, and a distorted sense of reality (as noted in Similar Post #2). However, when the manic phase subsides, the emotional and physical toll can be overwhelming. This "comedown" may lead to a **mixed episode** (a blend of manic and depressive symptoms) or a rapid shift into depression. Key factors that might contribute to suicidal ideation during this phase include:  
1. **Guilt/Shame**: After impulsive or self-destructive actions during mania (e.g., financial ruin, relationship damage), individuals may feel intense guilt or shame.  
2. **Emotional Exhaustion**: The physical and mental strain of mania can leave individuals feeling drained and hopeless.  
3. **Hopelessness**: The cyclical nature of bipolar disorder can create a sense of futility, especially if treatment is inconsistent or ineffective.  
4. **Neurobiological Factors**: Fluctuations in brain chemistry (e.g., dopamine, serotonin) during mood transitions may heighten vulnerability to suicidal thoughts.  

### **Connecting to Class Themes**  
Your question ties into broader discussions in Chapter 14 about the **cyclical nature of bipolar disorder** (as highlighted in Similar Post #4) and the challenges of distinguishing it from other disorders like borderline personality disorder. Unlike borderline personality disorder, which involves unstable relationships and identity issues (Similar Post #3), bipolar disorder is marked by distinct mood episodes. However, the transition between these episodes can blur these lines, making diagnosis and treatment complex.  

### **Critical Thinking & Next Steps**  
- **Why might the post-manic phase be more dangerous than mania itself?** Consider how impulsivity during mania might lead to consequences (e.g., debt, isolation) that amplify despair later.  
- **How does this relate to treatment strategies?** Effective management of bipolar disorder often involves stabilizing mood swings (e.g., with medications like...
\end{lstlisting}

\section{DSPy Language Model Program for Discussion Forum Question Classification}
\label{app:dspy_question_classifier}

\lstset{
    language=Python,                      
    basicstyle=\ttfamily\small,
    keywordstyle=\color{blue}\bfseries,  
    stringstyle=\color{red},             
    commentstyle=\color{gray}\itshape,   
    numberstyle=\tiny\color{gray},
    backgroundcolor=\color{gray!10},     
    frame=single,                        
    breaklines=true,                    
    showstringspaces=false,
    tabsize=4
}

\begin{lstlisting}[language=Python, caption=DSPy Language Model Program for Zero-shot Discussion Forum Post Classification\newline]

class ClassifyDiscussionForum(dspy.Signature):
    """Classify the discussion forum post into one of the following categories based on the content of the forum post and guidelines from the instructor for the post (if applicable):
1. Academic Question - Questions about academic content posed to the teaching staff;
2. Academic Discussion - Usually a discussion forum post required by an assignment, or any type of discussion that does not contain an obvious question;
3. Logistics Question - Questions about course logistics;
4. Logistics Discussion - Other types of posts about course logistics;
5. Social - Discussion forum posts for social purposes."""
    post_content = dspy.InputField(desc="The content of the student's discussion forum post.")
    post_topic_content = dspy.InputField(desc="The guidelines from the instructor for this post.")
    post_classification = dspy.OutputField(desc="The classification for the post.")

\end{lstlisting}

\section{DSPy Language Model Program for the VTA Pedagogical Capability Classifier}
\label{sec:dspy_program_classifier}

\lstset{
    language=Python,                      
    basicstyle=\ttfamily\small,
    keywordstyle=\color{blue}\bfseries,  
    stringstyle=\color{red},             
    commentstyle=\color{gray}\itshape,   
    numberstyle=\tiny\color{gray},
    backgroundcolor=\color{gray!10},     
    frame=single,                        
    breaklines=true,                    
    showstringspaces=false,
    tabsize=4
}

\begin{lstlisting}[language=Python, caption=DSPy Language Model Program for Zero-shot Prompt-based Classifiers\newline]
from typing import List, Literal

class SimpleClassifier(dspy.Signature):
    """Given a discussion forum post and the corresponding teaching assistant response, classify how well this response adheres to pedagogical goals specified with a rubric. Provide your classification as integers."""
    
    discussion_forum_post = dspy.InputField(desc="The original student forum post.")
    ta_response = dspy.InputField(desc="The corresponding teaching assistant response.")
    pedagogical_rubric = dspy.InputField(desc="The pedagogical rubric you should follow when evaluating the response.")
    discussion_topic = dspy.InputField(desc="Description and instruction for the specific discussion topic for students.")
    course_info = dspy.InputField(desc="Information about the course.")
    rating: Literal[0, 1, 2, "NA"] = dspy.OutputField()


class ComplexSelfReflector(dspy.Module):
    def __init__(self, callbacks=None):
        super().__init__(callbacks)
        self.classifier = dspy.ChainOfThought(SimpleClassifier)
    
    def forward(self, discussion_forum_post, ta_response, pedagogical_rubric, discussion_topic, course_info):
        og_rating = self.classifier(discussion_forum_post=discussion_forum_post,
                                    ta_response=ta_response,
                                    pedagogical_rubric=pedagogical_rubric,
                                    discussion_topic=discussion_topic,
                                    course_info=course_info).rating
        return dspy.Prediction(rating=og_rating)
\end{lstlisting}

\twocolumn

\begin{table*}[]
    \centering
    \caption{Full rubric for evaluating VTA’s pedagogical capability in asynchronous online forums.}
    \begin{tabularx}{\textwidth}{p{2.3cm} X}
    \toprule
    \textbf{Level} & \textbf{Description}\\
    \midrule
    Clarify misunderstandings & 
    \begin{itemize}[leftmargin=*, nosep]
        \item \textbf{Strong (2):} Accurately identifies misunderstanding and confusion, provides a clear explanation using relevant content and examples.
        \item \textbf{Weak (1):} Attempts to address the question or confusion, but the explanation is vague or lacks instructional clarity.
        \item \textbf{Not Present (0):} Fails to recognize or respond to the misunderstanding or question.
        \item \textbf{Not Applicable (NA):} No misunderstanding or question is present in the post.
    \end{itemize} \\
    \midrule
    Disciplinary Understanding &
    \begin{itemize}[leftmargin=*, nosep]
        \item \textbf{Strong (2):} Promotes deeper thinking and engagement with core disciplinary concepts by offering thought-provoking prompts, elaborations, or meaningful extensions.
        \item  \textbf{Weak (1):} Demonstrates intent to deepen disciplinary understanding, but through surface-level, or generic responses that lack meaningful connection to the post content.
        \item  \textbf{Not Present (0):} Makes no attempt to extend or deepen disciplinary understanding.
        \item \textbf{Not Applicable (NA):} Deepening disciplinary understanding is irrelevant to the post context.
    \end{itemize} \\
    \midrule
    Higher-Order Thinking &
    \begin{itemize}[leftmargin=*, nosep]
        \item \textbf{Strong (2):} Promotes higher-order thinking through specific, content-grounded prompts or reasoning tasks that challenge students to analyze, evaluate, or reflect.
        \item \textbf{Weak (1):} Attempts to promote higher-order thinking through general or loosely related prompts, but lacks depth, specificity, or clear alignment with the content.
        \item \textbf{Not Present (0):} Makes no attempt to encourage higher-order thinking.
        \item \textbf{Not Applicable (NA):} Higher-order thinking is not applicable given the context of the post.
    \end{itemize} \\
    \midrule
    Metacognitive Awareness &
    \begin{itemize}[leftmargin=*, nosep]
        \item \textbf{Strong (2):} Supports metacognitive awareness by using reflective prompts or strategies that help students assess their understanding, monitor their thinking, or make sense of their learning process in context.
        \item \textbf{Weak (1):} Encourages reflection or self-monitoring, but uses vague or generic language that is not tied to the student’s content or learning process.
        \item \textbf{Not Present (0):} Makes no attempt to promote metacognition or reflection.
        \item \textbf{Not Applicable (NA):} Metacognitive engagement is not relevant to the context of the post.
    \end{itemize} \\
    \midrule
    Collaborative Knowledge Construction &
    \begin{itemize}[leftmargin=*, nosep]
        \item \textbf{Strong (2):} Effectively fosters peer interaction by referencing specific student ideas, connecting diverse perspectives, or inviting further contributions in a personalized and contextually relevant manner.
        \item \textbf{Weak (1):} Demonstrates intent to encourage interaction, but relies on general prompts or surface-level invitations without engaging specific content or peer input.
        \item \textbf{Not Present (0):} Makes no attempt to promote peer interaction or build social connection.
        \item \textbf{Not Applicable (NA):}  Collaborative engagement is not relevant to the context of the student’s post.
        \end{itemize} \\
    \bottomrule
    \end{tabularx}
    \label{tab:rubric}
\end{table*}

\begin{table*}[]
    \centering
    \caption{Examples of responses at each rubric scoring level (Part 1).}
    \begin{tabularx}{\textwidth}{p{2.3cm} X}
    \toprule
    \textbf{Level} & \textbf{Student Posts and Response Examples}\\
    \midrule
    Clarify misunderstandings & 
    \textbf{Post:} Hi all, do you know what is the difference between opportunity cost and differential cost and how to distinguish them?

    \begin{itemize}[leftmargin=*, nosep]
        \item \textbf{Strong (2):} Great question! Opportunity cost is the value of the next best alternative you give up when making a choice. For example, if you spend an hour studying instead of working a \$20/hour job, the opportunity cost is \$20. Differential cost, on the other hand, refers to the difference in cost between two alternatives. If Option A costs \$500 and Option B costs \$700, the differential cost is \$200. So, opportunity cost is about forgone benefits, while differential cost is about financial differences between choices. Both help in decision-making, but they address different aspects.
        \item \textbf{Weak (1):} Opportunity cost is what you lose when you don’t choose another option, and differential cost is just the difference between two costs. They’re kind of related but not exactly the same. It’s sometimes hard to tell the difference, but maybe opportunity cost is more like what you’re missing out on, and differential cost is more like math. I guess you use both when comparing choices, but I’m not totally sure when each applies. Maybe they overlap in some situations, depending on what you’re deciding between.
        \item \textbf{Not Present (0):} Opportunity cost and differential cost both involve calculating how much something costs in total, so they can usually be used the same way. For example, if you’re choosing between two jobs, you’d just compare their salaries directly and include all previous expenses like tuition or commuting costs, even if you already paid them. In that case, sunk cost is also part of the decision. There’s not really a big difference between these costs—they all show up when comparing your options. You can just add them up to decide which one is better overall.
    \end{itemize} \\
    \midrule
    Disciplinary Understanding &
    \textbf{Post:} I’m trying to get a clearer understanding of circular satellite orbits—specifically, how their motion is described and how to calculate the total mechanical energy involved. In addition, I’m curious about how the concept of gravitational waves connects to phenomena like black holes. Could someone help explain these ideas in more detail?

    \begin{itemize}[leftmargin=*, nosep]
        \item \textbf{Strong (2):} In a circular orbit, the gravitational force equals the centripetal force, which gives the condition $v = \sqrt{\frac{GM}{r}}$. Using this, the mechanical energy (kinetic plus potential) comes out to $E = -\frac{GMm}{2r}$, the negative sign means the satellite is \textit{bound} to the planet. Notice how the farther the orbit (larger rrr), the closer the total energy moves toward zero, meaning the satellite is less tightly bound. Now compare that to black holes: as two orbit each other, they lose orbital energy by radiating gravitational waves, shrinking their separation until merging. Can you see the parallel?
        \item  \textbf{Weak (1):} A satellite in circular orbit moves because gravity pulls it in while its velocity keeps it moving around, so the two balance. Its energy comes from adding potential and kinetic energy together, which depend on mass and distance. Gravitational waves are waves in space caused by black holes. These topics are important in physics, and it could help to look at examples of how energy and gravity connect in different cases.
        \item  \textbf{Not Present (0):} Circular orbits, energy, and gravitational waves are all interesting parts of space science. Satellites go around planets, and black holes are mysterious objects that produce strange effects.These are important topics in physics.
    \end{itemize} \\
        \bottomrule
    \end{tabularx}
    \label{tab:example1}
\end{table*}

\begin{table*}[]
    \centering
    \caption{Examples of responses at each rubric scoring level (Part 2).}
    \begin{tabularx}{\textwidth}{p{2.3cm} X}
    \toprule
    \textbf{Level} & \textbf{Student Posts and Response Examples}\\
    \midrule
    Higher-Order Thinking &
    \textbf{Post:} The textbook noted that Social Security benefits were once fully excluded from income, but now up to 85\% may be taxed. Just wonder what led to this change?
    \begin{itemize}[leftmargin=*, nosep]
        \item \textbf{Strong (2):} That’s an insightful observation. The shift occurred mainly because of tax policy changes in the 1980s and 1990s aimed at strengthening Social Security’s long-term financing and addressing concerns about fairness in the tax system. Initially, Social Security benefits were fully excluded from taxation, but in 1983 Congress decided that up to 50\% of benefits could be taxed for higher-income recipients, and this was expanded to up to 85\% in 1993. Why do you think policymakers chose to tax only higher-income recipients instead of all beneficiaries? What does this reveal about the balance between revenue generation and protecting lower-income retirees?
        \item \textbf{Weak (1):} The change happened because Congress passed laws in the 1980s and 1990s that made Social Security benefits partly taxable. This helped raise revenue and keep the program more sustainable. How do you think this compares to other tax changes you’ve read about?.
        \item \textbf{Not Present (0):} Social Security benefits became partly taxable starting in 1983, with the share rising to as much as 85\% after 1993. This was the result of changes in federal tax law.
        \item \textbf{Not Applicable (NA):} Higher-order thinking is not applicable given the context of the post.
    \end{itemize} \\
    \midrule
    Metacognitive Awareness &
    \textbf{Post:} I’m confused about how bunching itemized deductions works. It seems to involve shifting deductions into one year to exceed the standard deduction, then using the standard deduction the next year. How exactly can a taxpayer shift deductions—what does it mean to “accelerate” payments into the current year?
    \begin{itemize}[leftmargin=*, nosep]
        \item \textbf{Strong (2):} Great question. The idea is that a taxpayer can “bunch” deductions by paying certain expenses earlier than usual—like making a January mortgage payment in December or prepaying charitable donations—so they fall into the same tax year. Can you walk through an example in your own words to check if this makes sense? For instance, how would prepaying property taxes affect whether you use the standard or itemized deduction this year versus next year?
        \item \textbf{Weak (1):} The taxpayer shifts deductions by paying some bills early, such as property taxes or donations, so they count in the current year. Think about how this changes the year’s total, and consider if it helps you see why the timing matters.
        \item \textbf{Not Present (0):} Taxpayers can shift deductions by paying certain expenses, like property taxes or charitable contributions, earlier so they fall into the same year and allow itemizing instead of taking the standard deduction.
        \end{itemize} \\
        \bottomrule
    \end{tabularx}
    \label{tab:example2}
\end{table*}

\begin{table*}[]
    \centering
    \caption{Examples of responses at each rubric scoring level (Part 3).}
    \begin{tabularx}{\textwidth}{p{2.3cm} X}
    \toprule
    \textbf{Level} & \textbf{Student Posts and Response Examples}\\
    \midrule
    Collaborative Knowledge Construction &
    \textbf{Post:} In the reading, the author explained that the main goal of sociology is to study the conditions that allow societies to continue existing. As I kept reading, I started to see how different kinds of societies are held together by different bonds and follow different sets of rules. What I’m still wondering, though, is \textit{why} this focus on conservation is so important?
    \begin{itemize}[leftmargin=*, nosep]
        \item \textbf{Strong (2):} I really like the way you phrased your question about “why” societies need to conserve themselves. It actually reminded me of \textit{Similar Post \#1}, where the author also wondered about the purpose of preserving societies in the first place. They suggested it might be about maintaining social order, and I think your point about rules connects well with that. I’m wondering if Durkheim saw conservation not just as survival, but as creating the conditions for people to cooperate and avoid chaos. Do you think his focus on mechanical vs. organic solidarity helps explain \textit{why} different societies rely on different rules to hold themselves together?
        \item \textbf{Weak (1):} That’s a good observation. I think the author was mostly interested in how societies stay stable over time, kind of like what some of the other posts mentioned. Maybe it’s just about survival. What do you think?
        \item \textbf{Not Present (0):} The author believed sociology’s goal was to identify the conditions that allow societies to survive, which is why different rules apply in different contexts.
        \end{itemize} \\
    \bottomrule
    \end{tabularx}
    \label{tab:example3}
\end{table*}

\onecolumn
\section{Additional VTA \cite{kweon2025large} Examples and Assessments}

\subsection{Example 1}
\includegraphics[page=1,scale=0.7]{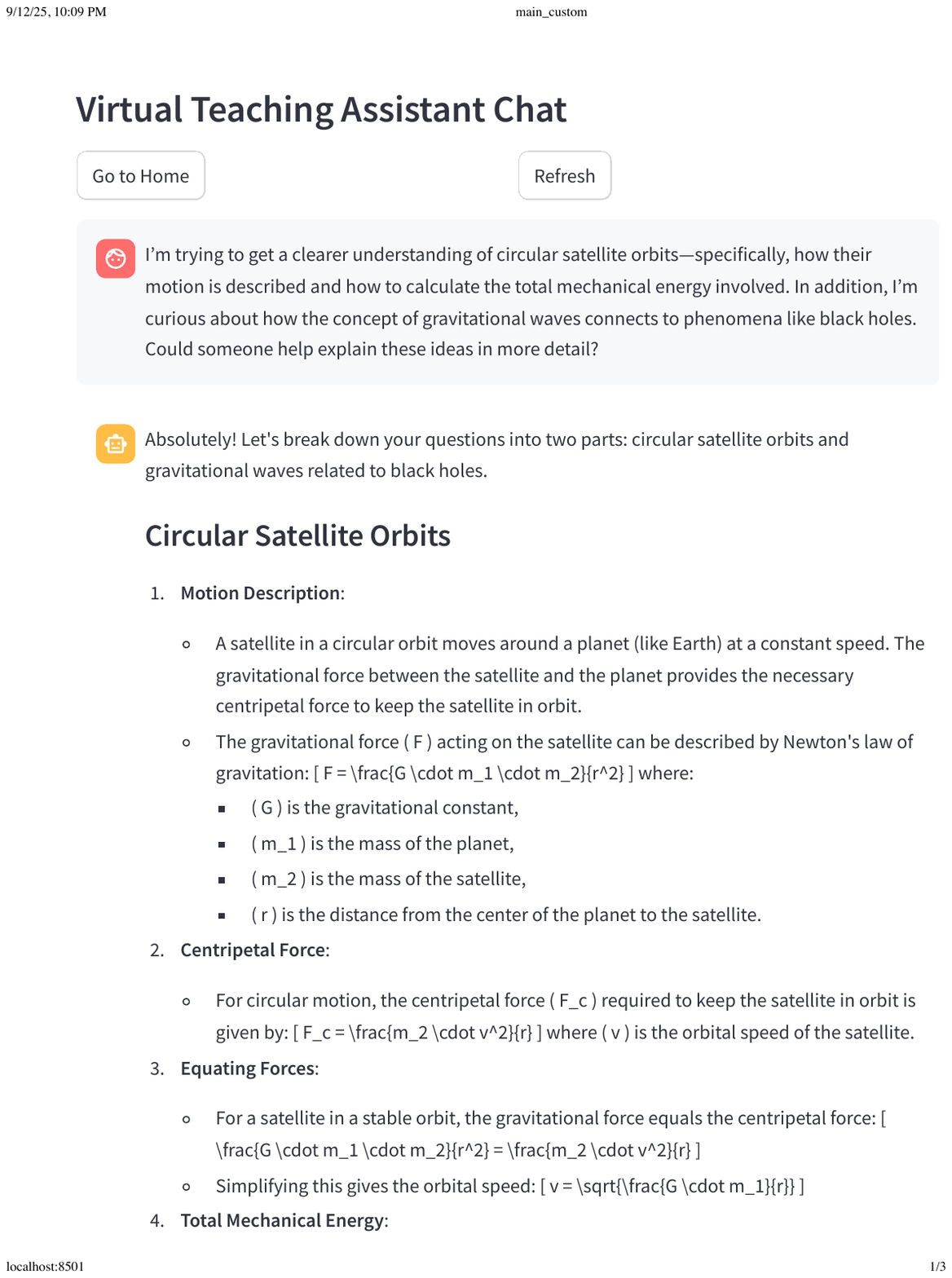}
\newpage
\includegraphics[page=2,scale=0.7]{figure/main_custom_2.pdf}
\newpage
\includegraphics[page=3,scale=0.5]{figure/main_custom_2.pdf}

\begin{enumerate}
    \item \textbf{Clarifying Misunderstandings and Confusions:} Weak (1)
    \item \textbf{Deepen Disciplinary Understanding:} Strong (2)
    \item \textbf{Foster Higher-Order Thinking:} Not Present (0)
    \item \textbf{Enhance Metacognitive Awareness:} Weak (1)
\end{enumerate}
\newpage

\subsection{Example 2}
\includegraphics[page=1,scale=0.7]{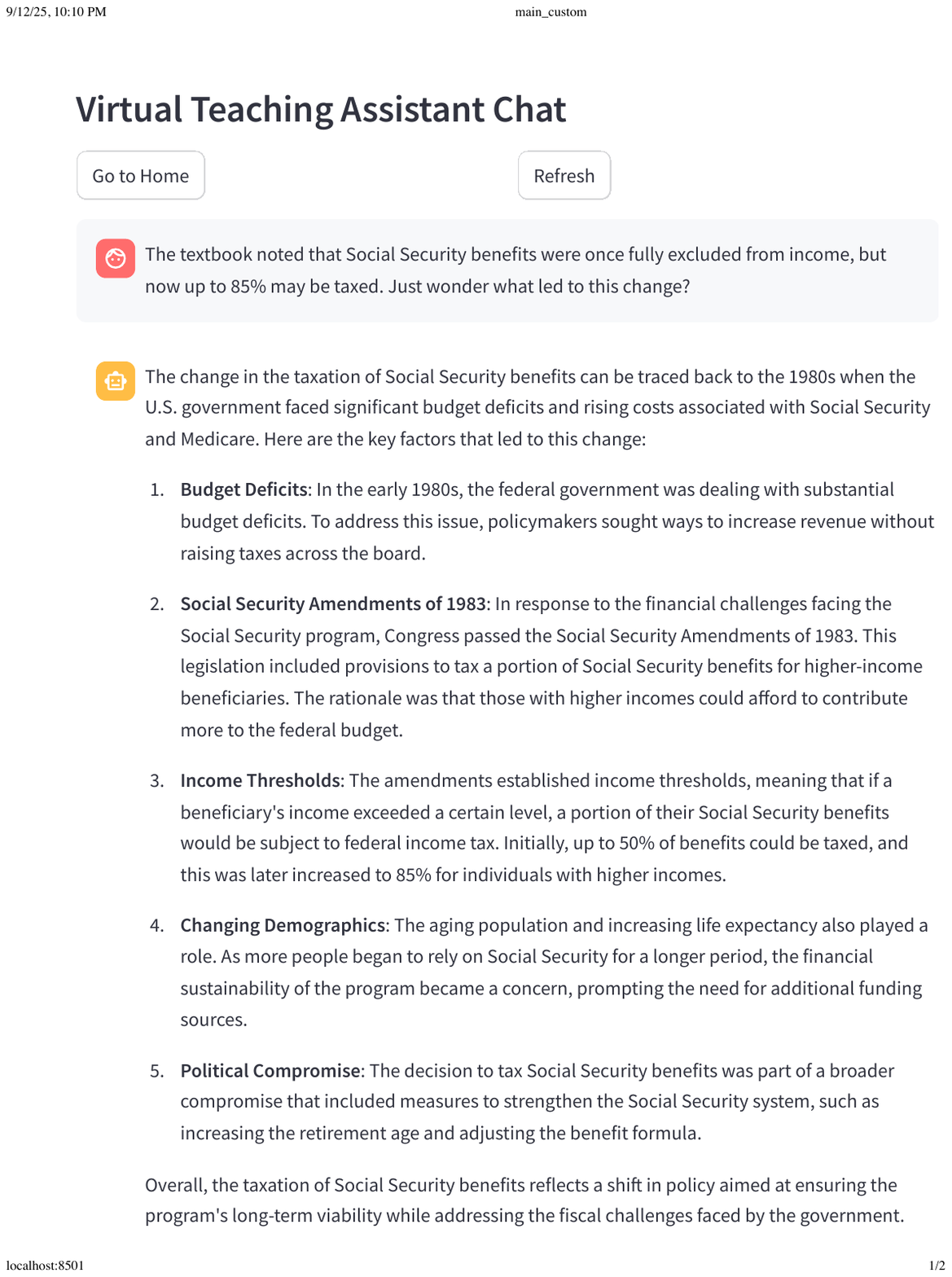}

\begin{enumerate}
    \item \textbf{Clarifying Misunderstandings and Confusions:} 2
    \item \textbf{Deepen Disciplinary Understanding:} 2
    \item \textbf{Foster Higher-Order Thinking:} 0
    \item \textbf{Enhance Metacognitive Awareness:} 0
\end{enumerate}
\newpage

\subsection{Example 3}
\includegraphics[page=1,scale=0.7]{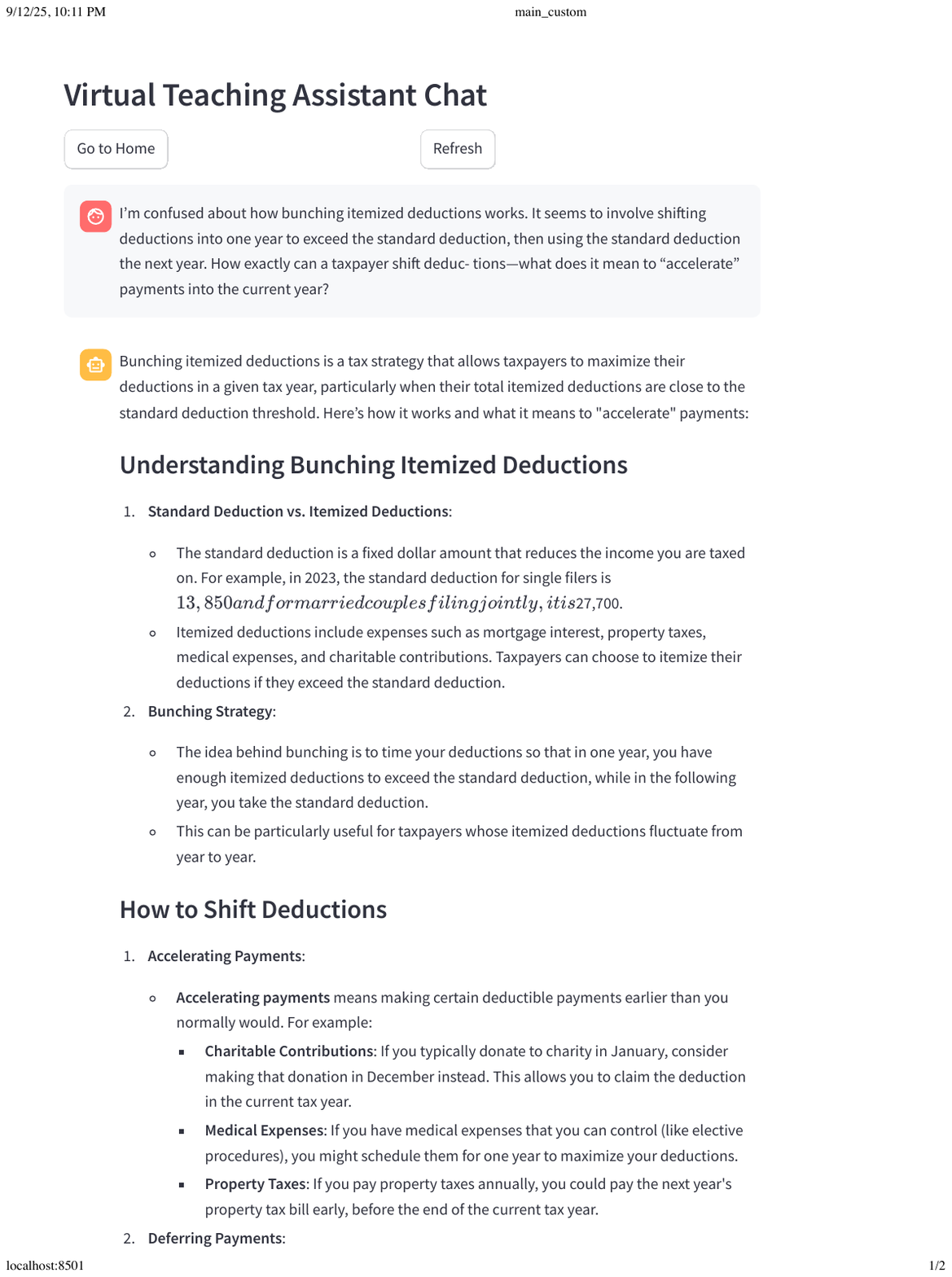}
\newpage
\includegraphics[page=2,scale=0.5]{figure/main_custom_4.pdf}

\begin{enumerate}
    \item \textbf{Clarifying Misunderstandings and Confusions:} 2
    \item \textbf{Deepen Disciplinary Understanding:} 1
    \item \textbf{Foster Higher-Order Thinking:} 0
    \item \textbf{Enhance Metacognitive Awareness:} 0
\end{enumerate}

\end{document}